\pdfoutput=1

\documentclass[11pt]{article}

\usepackage{authblk}

\usepackage{EMNLP2022}

\usepackage{times}
\usepackage{latexsym}

\usepackage[T1]{fontenc}

\usepackage[utf8]{inputenc}

\usepackage{microtype}

\usepackage{inconsolata}

\usepackage{lscape}
\usepackage{spreadtab}
\usepackage{graphicx}
\usepackage{color,soul}
\usepackage{siunitx}

\title{Human-in-the-Loop Hate Speech Classification in a Multilingual Context}

\author[]{\textbf{Ana Kotarcic}\textsuperscript{1,2}, \textbf{Dominik Hangartner}\textsuperscript{2}, \textbf{Fabrizio Gilardi}\textsuperscript{1}, \\ 
\textbf{Selina Kurer}\textsuperscript{2} and \textbf{Karsten Donnay}}

\affil[1]{Department of Political Science, University of Zurich, Switzerland}
\affil[2]{Immigration Policy Lab, ETH Zurich, Switzerland}

\affil[ ]{ana.kotarcic@gmail.com,
dominik.hangartner@gess.ethz.ch,
gilardi@ipz.uzh.ch,
}
\affil[ ]{
selina.kurer@gess.ethz.ch,
donnay@ipz.uzh.ch
}

\begin{document}
\maketitle
\begin{abstract}
The shift of public debate to the digital sphere has been accompanied by a rise in online hate speech. While many promising approaches for hate speech classification have been proposed, studies often focus only on a single language, usually English, and do not address three key concerns: post-deployment performance, classifier maintenance and infrastructural limitations. In this paper, we introduce a new human-in-the-loop BERT-based hate speech classification pipeline and trace its development from initial data collection and annotation all the way to post-deployment. Our classifier, trained using data from our original corpus of over 422k examples, is specifically developed for the inherently multilingual setting of Switzerland and outperforms with its F1 score of 80.5 the currently best-performing BERT-based multilingual classifier by 5.8 F1 points in German and 3.6 F1 points in French. Our systematic evaluations over a 12-month period further highlight the vital importance of continuous, human-in-the-loop classifier maintenance to ensure robust hate speech classification post-deployment. 

\end{abstract}

\section{Introduction}
Hate speech, often taken to designate insults and attacks against individuals or groups based on their inherent traits, is an expression which is widely used but on whose definition there is no general consensus. Given the subjective and contextual nature of hate speech, it is left open to media platforms \cite{fb-hs-def, twitter-hs-def, youtube-hs-def, microsoft-hs-def}, research groups \cite{djuric:2015, saleem:2017, mondal:2017, salminenetal:2018, jaki-desmedt:2019, pereira-kohatsu:2019, rani-2020, roettger:2021} and individuals to decide what to include under this notion and whether to adjust its definition \cite{fb_adapted_hs_def:2022}. Particularly difficult is drawing the line between hate speech, toxic speech and humour expressed through e.g. irony, sarcasm or euphemisms.

The challenges described in this paper show that configuring a stable and robust hate speech classifier is not a trivial task. While this applies to big tech companies who have the data and infrastructure to create, train and maintain their own classification systems, it is often prohibitive for smaller, mostly regional, companies and (online) media outlets. For them, no tailored off-the-shelf solution exists, they often do not have the in-house capacity to develop and maintain their own classification system and are bound by data protection rules tightly regulating which data can be shared with commercial services. The Swiss context presents a particularly challenging case given that many instances of hate speech are deeply ingrained in the multilingual, sociocultural and political contexts of Switzerland \citep[cf. e.g.][]{stevenson:1990, hega:2001, bruegger:1990, mueller:2014, fdfa:2020} and are not picked up by currently available toxic speech classifiers. This situation is exacerbated by the fact that, up to the point of completing this study, no annotated hate speech data was available on which a solid baseline classifier for the Swiss context could have been trained.

The classification framework we developed addresses an important research gap regarding long-term, stable hate speech classification and is directly motivated by how such a classifier would be deployed in practice. A core feature of this system is that the human analysts remain both in control and at the centre of the decision making process, which we believe is vital to ensuring good post-deployment performance and enabling classifier maintenance. In building our human-in-the-loop hate speech classification pipeline, we closely collaborated with several Swiss media outlets and a Swiss NGO working on countering online hate.

In what follows, we will describe in more detail the steps involved in constructing this pipeline from the data collection stage all the way to post-deployment testing and the challenges encountered in the process. Due to non-disclosure agreements signed with our partners, we are not able to disseminate our dataset or code, but we can share the technical details of our framework and approach. We also include an extensive appendix describing the nature of our dataset and details relating to the annotation process.

\section{Related Work}
Research on online hate speech and its detection has risen exponentially over the past few years, yielding a wide variety of approaches to tackling the phenomenon. Given the large body of work, numerous surveys and overviews outlining the main directions of research have been published on the subject \cite{schmidt-wiegand:2017, salminenetal:2018,fortuna:2019,vidgen:2019, abro:2020, macavaney:2019, siegel:2020, mohiyaddeen:2021, yin:2021, mullah:2021, tontodimamma:2021}. The research they review has mostly focused on optimising classifier performance by focusing on different types of input data, preprocessing steps, languages, classifier types, model architectures and feature engineering options, and by performing classification on various types of problematic language and target groups (cf. \hyperref[table:Table 8]{Table 8} in Appendix B). 

Additionally, a series of shared tasks on offensive, abusive and hate speech classification have featured in NLP and linguistics competitions (cf. \hyperref[table:Table 9]{Table 9} in Appendix B) and are usually carried out in controlled environments with carefully configured datasets. Tasks are typically divided into two main categories: binary classification and multiclass classification, both for specific target groups and different types of offensive language.

The results of these challenges highlight that both the model selection and the nature of the dataset have a direct influence on classification performance whereby transformer models \cite{vaswani:2017}, particularly those based on the BERT architecture \cite{devlin:2019}, significantly outperform other models. Multiclass classification approaches perform consistently worse than do binary ones. When more than one language is present in the dataset, classification is usually not carried out in a multilingual setting but for each language individually. Classifiers for English often perform much better due to the sheer number of data points and pretrained models available, reaching F1 scores of 83.0 for binary and 66.6 for multiclass classification \cite{mandletal_2021:2021}. In contrast, the best-performing German language classifiers reach F1 scores of 77.0 for binary and 54.9 for multiclass classification, which can partly be attributed to the complex structure of the German language \cite{corazza:2020}. Outside of these challenges, the thus-far best-performing BERT-based multilingual classifier reaches an F1 score of 74.7 for German and 76.9 for French \cite{deshpande:2022}.

Despite this proliferation of research on hate speech, the topic has, to date, received very limited scholarly attention in the Swiss context, a particularly complex linguistic, sociocultural and political landscape. A notable exception is the work by \citet{binder:2020}, who explore gendered hate speech in Swiss WhatsApp messages. 

Research on post-deployment performance, classifier maintenance and infrastructural limitations is similarly scarce. Some attempts at discussing post-deployment performance in terms of cross-domain and concept drifts have been made \cite{vidgen:2019, sood:2012, aljero:2021, zhang:2018, malik:2022, ribeiro:2020, salminen:2020, bosco:2018, karan-snajder:2018} and show that performances of deployed models are often lower than those obtained \textit{in vitro}, sometimes dropping as much as 15 F1 points \cite{arango:2019}. While these insights are valuable, all of these studies only artificially post-test models with publicly available datasets: none of them presents long-term pre- and post-deployment results. 

Similarly, infrastructural challenges encountered in the process of training, evaluating and testing remain understudied. \citet{salminen:2021} provide the currently most complete examination of challenges related to automatic hate speech detection. Of the 19 action points Salminen et al. describe, the setup of our classification pipeline allows us to contribute to addressing the following five: AP02 (developing a multilingual hate speech classifier), AP03 (acquiring new samples periodically to refresh the training sets used for model development), AP13 (collecting and testing datasets from multiple social media platforms), AP18 (measuring and displaying overall error rates of online hate classifiers to end-users such as moderators) and AP19 (providing probabilities and explanations of labels provided by the hate classifier for individual samples).

\section{Methodology}
\subsection{Definition of Hate Speech}
For this study, we opted for a combination of top-down and bottom-up approaches to establishing an onion-like definition of hate speech. At its heart lies the UN statement that hate speech is ``any kind of communication in speech, writing or behaviour, that attacks or uses pejorative or discriminatory language with reference to a person or a group on the basis of who they are, in other words, based on their religion, ethnicity, nationality, race, colour, descent, gender or other identity factor'' \citep[p. 2]{UN-HS-def}. Additionally, we identified further characteristics and groups which were often targeted in online discourse in Switzerland. We therefore expanded our definition of hate speech to reflect those. Our final operational definition of hate speech comprises attacks and insults against the following target groups:

\begin{itemize}
    \item sex, age, gender, religion, nationality/skin colour/origin, and mental and bodily impairments (UN definition)
    \item social status (e.g. income, education, job), political orientation and appearance
    \item other (e.g. Covid-19, cyberbullying)
\end{itemize}

In practice, it is not straightforward to delimit toxic speech, i.e. insults and derogatory use of language, from statements targeting a specific person or group based on their (inherent) characteristics. Many toxic claims linguistically closely resemble hate speech statements. Therefore, instances of toxic speech were also annotated as positive examples. In so doing, we could systematically evaluate how our classifier configured for hate speech performs in comparison to one fine-tuned on a looser, toxicity-based definition which corresponds more closely to many of the classification targets used in research to date.

\subsection{Human-in-the-loop Hate Speech Classification Pipeline}
Our objective was to build a binary, multilingual, human-centered, long-term sustainable and self-sufficient classification pipeline which enables robust identification of (online) hate speech and allows our NGO and media partners to react to hate speech in a more timely fashion. So far, classification pipelines often end after prediction has taken place \cite{abro:2020,aljero:2021, mullah:2021, pereira-kohatsu:2019}. While this is reasonable in research settings, it does not allow the pipeline to remain up-to-date long-term, given the constant flux in language usage, style, discourse and topics of discussion \cite{florio:2021}. To bypass this issue, we propose incorporating an automatic retraining process into the pipeline, similar to that suggested by \citet{alsafari:2021}, with the exception that our retraining involves a human-machine hybrid decision making process as part of the active learning and refining framework stages \cite{budd:2019, enarsson:2022}. 

In this setup, two sets of labels are produced: weak labels, which are those generated by the classifier prior to human checks, and strong labels, which are those our annotators confirmed to be (non-)hate speech. In practice this means that an initially trained classifier is used to classify new data. The resulting weak labels are then checked by human coders who confirm or reject them and annotate the target groups. Once checked, the annotated examples -- drawn from newspaper comments or tweets -- are added to the existing training set before the model is automatically retrained either after a certain period of time has passed or a certain volume of new content is available. The progressive accumulation of annotated data and knowledge about hate speech allows the model to learn incrementally and adapt to changes, thus yielding more robust long-term results. To ensure that the model does not overfit and that parameters remain up-to-date, checks of the training results are carried out with parameter tuning taking place when and where necessary. The latter can be facilitated by a built-in function which, when called, launches a define-by-run API provided by the open-source optimisation software OPTUNA \cite{akiba:2019} and automatically searches for the best hyperparameters using a search-and-prune algorithm for cost-effective optimisation.

\subsection{Data Set and Data Set Balancing }
All datasets stem from the Swiss Hate Speech Corpus specifically compiled for this study. The corpus, which is described in \hyperref[section:Annotation Process]{Appendix A} in more detail, was annotated by a total of 19 trained student research assistants as well as volunteers recruited through our NGO partner. Annotators were instructed first to decide whether or not a given example features hate speech or toxic speech. In cases where an example was hate speech, annotators were instructed to carry out multilabel annotation of the groups targeted. In cases where no target group was recognisable, annotators were asked to label the example as toxic speech. Most examples were annotated only a single time. Quality control was ensured by measuring the inter-rater-reliability for a subset of examples which were included in each annotation set, by performing qualitative checks and by holding bi-weekly deliberation rounds with annotators to clarify unclear cases. 

The latter was particularly important given both the difficulty of the task and the heterogeneous nature of the data. While we focused on examples written in  (Swiss) German and French, our data also includes numerous examples which feature terms in other national and frequently used languages in Switzerland, different Swiss German dialects as well as various writing scripts (e.g. Arabic, Cyrillic, Greek) and reading directions (e.g. Arabic). All of this introduced a host of linguistic and graphical issues which needed to be accounted for during the classification process. While other studies, like \citeposs{ousidhoum:2019}, have excluded examples featuring code switching from their training sets, this was not an option for us given the inherently multilingual and multi-dialectal context of Switzerland. 

With its over 422k unique annotations, this corpus is therefore one of the largest and linguistically most diverse corpora on hate speech \cite{poletto:2021}. The types and sources of data featuring in it were chosen in accordance with the requirements set by our NGO and media partners, without whom we would not have been able to carry out our project. As such, the corpus includes data from the following three main sources: (1) comments posted under online newspaper articles collected by our NGO partner (NGO); (2) online newspaper comments directly obtained from three Swiss national online newspaper outlets (ON1, ON2 and ON3); (3) tweets collected using the Twitter API of ``politically interested users'', i.e. accounts following at least five Swiss newspapers or politicians.

For our purposes, the most important set was the second, as it includes published, moderated and deleted comments from the three ONs, whose commentators -- taken together -- display varying linguistic, cultural and sociodemographic characteristics and are located all over Switzerland. To configure the classifier, we used all donated data, regardless of their moderation status. As expected from a corpus annotated for hate speech and the corresponding target groups \citep[cf.][]{tita:2021}, there are considerable class imbalances within the dataset \cite{geschke:2019}.

To mitigate these class imbalances and minimise overfitting, in the early stages of our work, we tested SMOTE, and numerous over- and undersampling strategies as well as different balancing ratios. Despite the distributional differences in the training and prediction sets, the best out-of-sample results were obtained by undersampling the majority class so as to obtain a 50-50 ratio of positive and negative examples in the training set. For the initial training set, all positively annotated instances were therefore joined with the same number of randomly sampled negatively annotated data points. For each retraining thereafter, the newly annotated data was balanced in such a way as to include an equal number of positive and negative examples, and was then added to the existing training set to produce an incrementally growing body of annotated examples. The same procedure was followed after deployment.

\subsection{Classification Models}
For training the classifier, we relied on two main architectures: multinomial Naïve Bayes (MNB), given its capability of performing text classification with few data points and in a computationally efficient manner, and on mBERT(\textsubscript{Base} uncased) \cite{devlin:2019, pires:2019, aluru:2020}, the thus-far most promising model for performing classification tasks in non-English languages due to its bidirectional nature and pre-training on 104 languages.
We also fine-tuned and tested a series of other BERT models using the HuggingFace open source library \cite{wolf:2019}: GELECTRA\textsubscript{Base} and GBERT\textsubscript{Base} \cite{chan:2020}, FlauBERT \cite{le:2020}, French RoBERTa\textsubscript{Base} \cite{french_roberta}, XLM-RoBERTa\textsubscript{Base} \cite{conneau:2019} and Twitter-XLM-RoBERTa\textsubscript{Base} \cite{barbieri:2021}. These models were chosen because they feature a wide variety of writing and language styles in their pretraining, which is important given the heterogeneous composition of our corpus; they also yield the most promising benchmark results for our type of task.

\subsection{Preprocessing}
Preprocessing steps for the MNB included the following: regularisation and stripping of extra white spaces, line breaks and multiple quotation marks, case folding, removal of punctuations, numbers, special characters, htmls and @-mentions, transcription of emojis to words, tokenisation, stopwords removal, lemmatisation. For the BERT-models, which are capable of grasping the semantics of sentences and taking into account word order, minimal preprocessing was carried out: regularisation, stripping of extra white spaces and line breaks, case folding, removal of htmls and @-mentions, transcription of emojis to words. For all steps, we used standard state-of-the-art libraries (nltk, re, emoji).

Transcribing emojis to words always means transcribing into English no matter the language of the rest of the input, which speaks in favour of using multilingual models. Moreover, we removed @-mentions given that our dataset displays some user names like \textit{@schwurbler} (slang for conspiracy theorist) which themselves are derogatory terms. Removing them helps to prevent our model from classifying potentially harmless examples as hate speech based simply on the user name.

\subsection{Training Parameters and Infrastructure}
\label{section:Training Infrastructure}
The bag-of-words MNB model was paired with tfidf and n-grams to ensure that at least some word order was preserved. A series of tests showed that best results are achieved with a maximum of 3000 tfidf features paired with a word n-gram range of 1-4 grams. The latter is based on \citet{malmasi:2017} and \citet{jauhiainen:2018}, who showed that 4-grams are most effective when processing (Swiss) German. Since our dataset also features numerous shorter words, including 1-3-grams further improved results. Model performance was evaluated using 10-fold-cross validation.

All transformer models were fine-tuned using the original model hyperparameters in order to allow for better comparisons between their performances. Leveraging data parallelisation, the models were trained for up to 5 epochs with a per-device batch size of 8 on a node consisting of 16 NVIDIA Tesla K80 GPUs with 16GB of system memory per GPU, running on average for approximately eight hours per training. The small batch size was conditioned on the fact that model parallelisation was not possible within our infrastructure, leading to out-of-memory (OOM) if a larger batch size was chosen. The latter constraint also meant that most attempts to perform hyperparameter tuning led to overfitting. Saving the best model ensured that the best performance was then used for final evaluation and prediction. All post-deployment performance evaluations were carried out on a single node consisting of 8 NVIDIA Tesla V100 GPUs with 32GB of system memory per GPU and a per-device batch size of 32, taking on average approximately four hours per prediction task. All GPUs ran with cuda 11.2 and cudnn 8.1.0. Models were trained and evaluated using a stratified, random train-test split with an 80-20 ratio.

\subsection{Deployment}
\label{section:Deployment}
For deployment, the classifier was integrated into a larger pipeline where new data in form of comments and tweets are periodically ingested from different Swiss online newspaper outlets and the Twitter API respectively. They are then passed through our human-in-the-loop hate speech classification pipeline which can be accessed by the NGO community and content moderators via a custom-built app hosted on university servers. In addition to presenting the comments and tweets in descending order of hate speech probability, the app allows users to perform checks of the hate speech labels and to carry out counter speech. Since all classification results are saved in our database, checked comments and tweets are automatically balanced and added to the training set for the next retraining round of the model.



\begin{table*}[h]
\tiny
\begin{center}
\noindent{\begin{center}\begin{tabular}{l l l l r l l r r r r r r}
\hline
\hline
& & \textbf{Train \& Eval} & & & \textbf{Pred} & & \textbf{MNB} & & & \textbf{mBERT\textsubscript{Base}} & & \\
\hline
\textbf{Wave} & \textbf{Week} & \textbf{Lang} & \textbf{Data} & \textbf{Size} & \textbf{Lang} & \textbf{Data} & \textbf{Precision} & \textbf{Recall} & \textbf{F1} & \textbf{Precision} & \textbf{Recall} & \textbf{F1}\\
\hline
\hline
1 & 1 & G & ON1 & 2.3k & G & ON1 & 62.2 & 62.3 & 62.2 & OF & OF & OF \\
& 2 & G & ON1 & 4.1k & G & ON1 & 65.2 & 65.2 & 65.2 & OF & OF & OF \\
& 3 & G & ON1 & 5.5k & G & ON1 & 63.2 & 63.1 & 63.0 & OF & OF & OF \\
& 4 & G & ON1 & 8.7k & G & ON1 & 65.0 & 64.9 & 64.9 & OF & OF & OF \\
& 5 & G & ON1 & 12.1k & G & ON1 & 64.0 & 64.0 & 64.0 & OF & OF & OF \\
& 6 & G & ON1 & 14.5k & G & ON1 & 67.4 & 67.4 & 67.3 & 75.9 & 75.8 & 75.8\\
& 7 & G & ON1 & 20.5k & G & ON1 & 65.4 & 65.3 & 65.3 & 73.6 & 73.3 & 73.2\\
& 8 & G & ON1 & 28.4k & G & Tweets & 66.6 & 66.5 & 66.4 & 73.3 & 72.7 & 72.5\\
& 9 & G & ON1 \& Tweets & 37.3k & G & Tweets & 67.1 & 66.9 & 66.8 & 72.3 & 72.3 & 72.2\\
\hline
\hline
2 & 1 & G & ON1 & 40.8k & FR & ON1 & 65.9 & 65.1 & 64.6 & 73.6 & 73.6 & 73.6\\
& 2 & G \& FR & ON1 & 47.6k & FR & ON1 & 73.0 & 69.2 & 67.8 & 77.7 & 77.5 & 77.4\\
& 3 & G \& FR & ON1 & 60k & FR & ON1 & 73.3 & 70.9 & 70.1 & 77.4 & 77.4 & 77.4\\
& 4 & G \& FR & ON1 & 67k & FR & ON1 & 71.9 & 70.4 & 69.8 & 77.3 & 77.1 & 77.1\\
& 5 & G \& FR & ON1 & 82.5k & FR & ON1 & 74.7 & 73.9 & 73.7 & 79.0 & 79.0 & 79.0\\
& 6 & G \& FR & ON1 & 94.5k & FR & ON1 & 75.2 & 74.8 & 74.7 & 79.3 & 79.3 & 79.2\\
& 7 & G \& FR & ON1 & 102.5k & FR & ON1 & 75.7 & 75.5 & 75.4 & 79.3 & 79.2 & 79.2\\
& 8 & G \& FR & ON1 \& Tweets & 119k & FR & Tweets & 74.2 & 73.8 & 73.7 & 79.1 & 79.0 & 79.0\\
& 9 & G \& FR & ON1 \& Tweets & 122k & FR & Tweets & 72.7 & 72.5 & 72.4 & 78.6 & 78.4 & 78.4\\
\hline
\hline
3 & 1 & G \& FR & ON1 \& Tweets & 127.7k & G & Tweets & 71.5 & 71.4 & 71.4 & 78.4 & 78.4 & 78.4\\
\hline
\hline
4 & 1-15 & G \& FR & ON1 \& Tweets & 144.6k & G & Tweets & 69.4 & 69.2 & 69.1 & 79.3 & 79.2 & 79.2\\
\hline
\hline
5 & 1-3 & G \& FR & ON1 \& Tweets & 177.9k & G & ON1 & 68.4 & 67.5 & 67.2 & 78.8 & 78.8 & 78.8\\
& 4-6 & G \& FR & ON1 \& Tweets & 186.7k & G & ON2 & 68.7 & 68.3 & 68.1 & 80.5 & 80.3 & 80.3\\
& 7-9 & G \& FR & ON1-2 \& Tweets & 191.6k & G & ON3 & 68.7 & 68.3 & 68.1 & 80.6 & 80.3 & 80.3\\
& 10-12 & G \& FR & ON1-3 \& Tweets & 197.4k & G & ONs \& Tweets & 69.3 & 68.8 & 68.6 & 80.8 & 80.5 & 80.5\\
\hline
\end{tabular}\end{center}}
\end{center}
\caption{Incremental Training Results}
\label{table:Table 1}
\end{table*}


\section{Results and Discussion}
\label{section:Results and Discussion}
To train and evaluate our classifier, we performed a series of experiments which align with the three stages of our development process: (1) incremental training which took place during the annotation process and where the classifier was retrained on a weekly basis; (2) pre-deployment experiments to cross-examine the final model results; and (3) post-deployment experiments where we tested the best-performing model \textit{in situ}. With the exception of one pre-deployment experiment, all pre-deployment training was carried out on data where occurrences of toxic speech were treated as positive examples. 

All experiments are reported using the weighted F1 score to account for the fact that post-deployment evaluation took place on a random subset of human-checked examples which the system had classified as hate speech. As such, in addition to human checks, F1 remained the most important performance metric throughout this study. Without any labeled hate speech data or baselines against which new results could be judged, for each retraining, a new F1 was generated which, depending on the result, may or may not have become the new baseline. In other words, the baseline against which we judged new results was always the best result which the model had achieved in the previous training round(s). This human-in-the-loop active learning approach proved to be a successful strategy as it allowed us concurrently to generate labeled data and configure the classifier. The final result reported in this study is that obtained upon the completion of our project. While we hope that it can serve as the baseline for future (Swiss) German and French hate speech classifiers, (re)training could have, in theory, continued until the F1 scores and the number of human-checked hate speech stagnate. 

Even though such a saturation point may eventually be reached \textit{in vitro}, it is unlikely that this will ever happen \textit{in situ}, given the shifting nature of language usage, topics of discussion and similar factors. Post-deployment retraining is therefore never-ending and is evaluated in the same way as during the annotation process. As such, F1 remains vital for post-deployment classifier maintenance. However, since the model is, upon deployment, fine-tuned for the (Swiss) German and French contexts, retraining would require fewer new data points -- just enough to introduce new features -- and would not have to be carried out as frequently as we needed to perform it, but certainly in moments of `dramatic' language shifts. As such, F1 can also be used for adjusting the retraining strategy and thus lowering carbon footprints, costs and not least human effort.

\subsection{Phase 1: Incremental Training}
\label{section:Incremental Training}

The first experiment aimed at systematically tracing how model performances change from one retraining to the next and at identifying potential pitfalls occurring during incremental training. The results obtained are listed in \hyperref[table:Table 1]{Table 1}, where ``G'' and ``FR'' indicate German and French, the source language of the data, respectively. The different data incrementation sizes reflect the number of annotated hate speech in any particular week and wave, and depended on the given stage of the annotation and retraining process, which model was used to generate weak labels, and on which language and data type we trained. 

Unlike the MNB, mBERT could not be used for (re-)training early on in the process. Only once a training set size of 14.5k data points was reached, was it possible to fine-tune the model without it overfitting (OF). This can partly be traced to the sources of examples and their respective distributions within the dataset, to the linguistic and graphical complexity of the training set and to the fact that the majority of examples were annotated only a single time. The latter in particular means the model has a much harder time performing its task, as was also suggested by \citet{leonardelli:2021}. They argue that the higher the per-sample inter-inter-rater reliability, the fewer samples are needed for the model to achieve good overall performance.

At the moment of introduction, mBERT instantly outperformed and continued to outperform the MNB, as is also visible in the annotation results (cf. \hyperref[table:Table 10]{Table 10} in Appendix B) which see a tripling in identified hate speech comments. In the process of retraining, mBERT, save for some minor oscillations, improved continually, confirming \citeposs{aluru:2020} observation that performances of classifiers improve with increasing number of data points on which the models are trained. The most pronounced oscillations happened at the point where prediction took place on French tweets: both the MNB and mBERT dropped in performance, but while mBERT was able to recover after a few retrainings, MNB performance continued to drop.

A different picture arose during zero-shot cross-lingual classification in wave 2 week 1, when the models -- fine-tuned on (Swiss) German comments only -- were used to make predictions on (Swiss) French comments which up to this point had not featured in the training set. Rather than causing a drop in performance, as was observed by e.g. \citet{pelicon:2021}, transfer learning between the two languages improved the F1 score by 1.4 points. Throughout the remainder of the iterative training, mBERT performance continued to improve, as is also reflected in the annotation results. For the MNB, the highest F1 score of 75.7 is recorded in wave 2 week 7; mBERT peaked in wave 5 week 12 with an F1 score of 80.5 and outperforms the thus-far best-performing BERT-based multilingual hate speech classifier \cite{deshpande:2022} by 5.8 F1 points in German and 3.6 F1 points in French.



\begin{table}[h]
\tiny
\begin{center}
\noindent{\begin{center}\begin{tabular}{l l r r r}
\hline
\hline
\textbf{Data} & \textbf{Toxic} & \textbf{Precision} & \textbf{Recall} & \textbf{F1}\\
\hline
\hline
G \& FR Comments \& Tweets & toxic as HS & 79.3 & 79.2 & 79.2\\
G \& FR Comments & no toxic & 80.0 & 79.9 & 79.8\\
G \& FR Comments & toxic as nonHS & 72.5 & 71.9 & 71.8\\
\hline
\end{tabular}\end{center}}
\end{center}
\caption{Model Performance according to Toxic Speech in the Training Set}
\label{table:Table 2}
\end{table}




\begin{table*}[h]
\tiny
\begin{center}
\noindent{\begin{center}\begin{tabular}{l l l r r r r r r}
\hline
\hline
& & & \textbf{Eval} & & & \textbf{Test} & & \\
\hline
\textbf{Train \& Eval Data} & \textbf{Train \& Eval Platform} & \textbf{Test Platform} & \textbf{Precision} & \textbf{Recall} & \textbf{F1} & \textbf{Precision} & \textbf{Recall} & \textbf{F1}\\
\hline
\hline
G \& FR Comments & ON1 & ON1 & 79.8 & 79.6 & 79.6 & 78.8 & 78.8 & 78.8\\
G \& FR Comments & ON1 & Twitter & 79.8 & 79.6 & 79.6 & 66.3 & 71.6 & 65.4\\
\hline
G \& FR Tweets & Twitter & ON1 & 76.4 & 76.4 & 76.4 & 68.7 & 68.1 & 68.3\\
G \& FR Tweets & Twitter & Twitter & 76.4 & 76.4 & 76.4 & 70.7 & 70.1 & 70.7\\
\hline
G \& FR Comments \& Tweets & ON1 \& Twitter & ON1 & 79.3 & 79.2 & 79.2 & 75.8 & 74.3 & 74.9\\
G \& FR Comments \& Tweets & ON1 \& Twitter & Twitter & 79.3 & 79.2 & 79.2 & 70.6 & 70.1 & 69.4\\
\hline
\end{tabular}\end{center}}
\end{center}
\caption{Cross-Platform Classification Results}
\label{table:Table 3}
\end{table*}




\begin{table*}[h]
\tiny
\begin{center}
\noindent{\begin{center}\begin{tabular}{l l l r r r r r r}
\hline
\hline
& & & \textbf{Eval} & & & \textbf{Test} & & \\
\hline
\textbf{Data} & \textbf{Train \& Eval Data} & \textbf{Test Data} & \textbf{Precision} & \textbf{Recall} & \textbf{F1} & \textbf{Precision} & \textbf{Recall} & \textbf{F1}\\
\hline
\hline
G \& FR Comments & ON1 & ON1 & 79.8 & 79.6 & 79.6 & 78.8 & 78.8 & 78.8\\
G \& FR Comments & ON1 & ON2 & 79.8 & 79.6 & 79.6 & 77.9 & 82.3 & 75.5\\
G \& FR Comments & ON1 & ON3 & 79.8 & 79.6 & 79.6 & 92.2 & 96.0 & 94.1\\
G \& FR Comments & ON1 + ON2 & ON3 & 80.6 & 80.3 & 80.3 & 92.2 & 96.0 & 94.1\\
G \& FR Comments & ON1 + ON3 & ON2 & 80.8 & 80.7 & 80.6 & 75.0 & 82.0 & 75.0\\
\hline
\end{tabular}\end{center}}
\end{center}
\caption{Cross-Dataset Classification Results}
\label{table:Table 4}
\end{table*}

\subsection{Phase 2: Pre-Deployment}
\label{section:Pre-Deployment}
The second experiment aimed at establishing whether mBERT was indeed the best BERT model for our purposes. For this, we trained and evaluated a series of other models whose performance results are detailed in \hyperref[table:Table 11]{Table 11} in Appendix B. For each model and dataset, the transformer model in question was first used for prediction (base) before it was fine-tuned and evaluated on our annotated dataset (tuned). Despite expecting, and getting, very poor results during the base runs, it emerged that model performance already at this stage is language and data type dependent. On average, GELECTRA performed best for German comments, and German comments and tweets, GBERT for German tweets, flauBERT for all French data types, XLM-RoBERTa and Twitter-XLM-RoBERTa for multilingual comments and XLM-RoBERTa for multilingual tweets.

This is a valuable observation given that implicit biases are present in these models and remain present even after fine-tuning. We noticed this propagated bias in wave 1 week 8 of the annotation process when the majority of tweets in our dataset still stemmed from Germany -- we later adapted our filtering process to minimise this phenomenon. Even though our classifier was capable of picking up some of the hate speech tweets with a distinctly Swiss German context, mBERT was much more ready and capable to pick up hate speech in tweets written in German German. This is presumably because most of the German speaking Wikipedia texts on which the model was pretrained will have stemmed from a German context.

Once fine-tuned, the models yielded varying performances. Models fine-tuned and evaluated exclusively on (Swiss) German comments yielded similar results, whereby the monolingual outperformed the multilingual models. In French, flauBERT outperformed french RoBERTa, presumably because of the varied nature of the datasets on which it was pretrained. No results can be reported on French tweets given the small size of the training set which led to almost instantaneous overfitting. For all data types, the multilingual models performed better on the pure French sets than did the monolingual models, suggesting that monolingual models for French are not yet as accurate as their German counter parts. While the multilingual models performed similarly when trained and evaluated on multilingual datasets, mBERT significantly outperformed XLM-RoBERTa and Twitter-XLM-RoBERTa once both comments and tweets were used. We therefore proceeded to deploy this model.

However, before final deployment, we carried out another experiment to examine whether or not and, if so, how toxic speech should be incorporated in the training set. Throughout the annotation process, occurrences of toxic speech were annotated as positive examples. Since our classifier is meant primarily to look for hate speech not toxic speech in the wider sense, three tests were carried out: in the first, examples featuring toxic speech remained under the positive label, as they did in all other experiments; in the second, they were dropped from the training set; in the third, examples of toxic speech were included in the training set as negative examples.

The results of these experiments are summarised in \hyperref[table:Table 2]{Table 2}. They show that reclassifying toxic speech to the negative label significantly decreased performance. Leaving toxic speech under the positive label or dropping toxic speech entirely yielded similar quantitative performances. Thereby the classifier without toxic speech performed slightly better. A qualitative analysis confirmed that training the classifier without toxic speech sharpens the focus, leading to fewer toxic examples and more hate speech proper, i.e. attacks against specific target groups, being recognised.

Determining these differences was only possible by inspecting the predictions and manually comparing their results. The evaluation metrics only marginally helped, suggesting that in the future, in settings such as ours, evaluation metrics are needed which also account for qualitative performance differences. Given these results, the deployed mBERT classifier does not feature toxic speech in the training set.

\subsection{Phase 3: Post-Deployment}
\label{section:Post-Deployment}

To test the robustness and long-term sustainability of the classifier, further experiments were carried out post deployment. Since post-deployment evaluation is identical to the active learning and incremental pre-deployment evaluation which took place during the annotation process, all annotations resulting from these tests were added to the Swiss Hate Speech Corpus (wave 5). This manner of proceeding allowed for further tracing of the incremental training results post-deployment and served as a stress-test for the hate speech classification pipeline when used \textit{in situ}. 

In the first experiment, we trained and evaluated the model on German and French ON1 comments, then on tweets and finally on ON1 comments and tweets, all taken from the first four annotation waves, before using the deployed mBERT and the MNB for comparison to predict on data from ON1 and Twitter. As the results in \hyperref[table:Table 3]{Table 3} show, zero-shot cross-platform classification significantly decreased model performance with training on ON1 and prediction on Twitter causing the largest drop of 14.2 F1 points.



\begin{table*}[h]
\tiny
\begin{center}
\noindent{\begin{center}\begin{tabular}{l r r r r r r r r}
\hline
\hline
& & & \textbf{Eval} & & & \textbf{Test} & & \\
\hline
\textbf{Train Data} & \textbf{Train and Eval Period} & \textbf{Testing Period} & \textbf{Precision} & \textbf{Recall} & \textbf{F1} & \textbf{Precision} & \textbf{Recall} & \textbf{F1}\\
\hline
\hline
G \& FR Comments & 01-2019 to 06-2019 & 07-2019 to 07-2021 & 74.6 & 74.5 & 74.5 & 78.2 & 51.5 & 51.5\\
G \& FR Comments & 01-2019 to 12-2019 & 01-2020 to 07-2021 & 80.1 & 79.6 & 79.5 & 58.9 & 62.5 & 60.0\\
G \& FR Comments & 01-2019 to 06-2020 & 07-2020 to 07-2021 & 80.0 & 79.6 & 79.5 & 83.9 & 75.8 & 79.0\\
G \& FR Comments & 01-2019 to 12-2020 & 01-2021 to 07-2021 & 79.9 & 79.7 & 79.6 & 81.6 & 74.6 & 77.2\\
G \& FR Comments & 01-2019 to 06-2021 & 07-2021 to 07-2021 & 79.2 & 79.1 & 79.1 & 82.1 & 76.8 & 78.9\\
G \& FR Comments & 01-2019 to 07-2021 & 01-2022 to 03-2022 & 80.0 & 79.9 & 79.8 & 65.1 & 71.5 & 65.5\\
\hline
\end{tabular}\end{center}}
\end{center}
\caption{Temporal Drift}
\label{table:Table 5}
\end{table*}


\begin{table}[h]
\tiny
\noindent{\begin{center}\begin{tabular}{r r r r r}
\hline
\textbf{HS Probability} & \textbf{Total} & \textbf{HS} & \textbf{nonHS} & \textbf{HS\%}\\
\hline
\hline
1.00-0.90 & 3995 & 3601 & 394 & 90\\
0.89-0.85 & 3952 & 2946 & 1006 & 75\\
0.89-0.80 & 7717 & 5458 & 2259 & 71\\
0.79-0.70 & 7952 & 4197 & 3755 & 53\\
0.69-0-60 & 7951 & 3274 & 4677 & 41\\
0.59-0.50 & 6207 & 1897 & 4310 & 31\\
\hline
\end{tabular}\end{center}}
\caption{Thresholding}
\label{table:Table 6}
\end{table}


Similar results emerged from the second experiment in which we performed cross-dataset testing on ON1, ON2 and ON3. To ensure the model would not overfit due to small numbers of training samples, ON1 comments annotated throughout the first four waves featured in each of the training sets. Prediction was performed on a random sample of the ON which the model had not yet seen. As \hyperref[table:Table 4]{Table 4} shows, performance dropped for all ONs up to 5.6 F1 points except for ON3 where we obtained significantly higher results than during evaluation. This can be explained by the fact that ON3, unlike ONs 1 and 2, features clearer cases of hate speech.

In the final experiment, we tested for temporal drift by using different temporal cut-offs to create various combinations of training and test sets. The results are summarised in \hyperref[table:Table 5]{Table 5} and suggest that the more offset in time the prediction set is, the less accurate the classifier becomes. This is particularly visible in the last test where the training and prediction sets are almost 6 months apart. The result of this time shift was a drop in performance by 14.3 F1 points.

To mitigate some of this classifier drift and provide more robustness to the deployed model, we conducted an analysis of how hate speech samples behave according to predicted hate speech probability. In practice, this allows us to provide empirically-grounded recommendations for which hate speech threshold end users should use when deploying our model. As shown in \hyperref[table:Table 6]{Table 6}, the number of hate speech examples decreases the further down the probability scale we go. This was also reflected in qualitative analyses which showed that the further down the probability scale we go, the more contested and borderline cases we find. 

Striking in our case is the sharp decline in the number of hate speech tweets in the 0.89-0.80 probabilities band when compared to that between 1.00 to 0.90. For our deployed model, this indicates that if thresholding were to be applied, the optimal cutoff point would be at 0.90. Depending on how much tolerance is acceptable, the threshold could be lowered: yet to guarantee the model performance obtained at the time of training and evaluation, the threshold should be set no lower than 0.85, at which point 80 percent of positively classified examples will still be hate speech. Any threshold below 0.85 would no longer guarantee optimal model performance.

\section{Conclusion}
\label{section:Discussion and Conclusions}
This paper presents a multilingual human-in-the-loop BERT-based hate speech classification pipeline adapted to the Swiss context. By systematically tracing and reporting classifier performances from the start of the annotation process to post-deployment, we highlight important practical considerations for fine-tuning a state-of-the-art BERT-based classification model. The most important is that testing classifiers post-deployment is vital for configuring robust systems, as various types of drift cause performances to drop. Next to focusing on ameliorating systems pre-deployment, future research should thus also look into possible mitigation techniques to stabilise classifier performance post-deployment.

In addition to recommending thresholds, a first step in that direction is to discard the idea of a static, one-time-only-trained model in favour of an adaptive learning concept \cite{gama:2014, laaksonen:2020}. In this paradigm, the classifier functions like a living organism and evolves together with the ever-changing uses of language, and alterations in mindsets and topics. This can only be achieved by actively maintaining the classifier post-deployment, i.e. by dynamically adjusting the definition of hate speech and the annotation schemes, by periodically retraining the model and by allowing it progressively to accumulate and adapt its knowledge of hate speech.

\section{Limitations}
\label{section:Limitations}
The biggest issues we faced throughout this study were infrastructural limitations related to scalability. Due to NDAs preventing us to upload our raw newspaper comment data to cloud services and similar infrastructures, all training, evaluation and testing was carried out on the university cluster. With each incremental increase in training set size, more memory and time was needed for retraining the model. Restrictions on the availability of GPU memory in particular meant that, despite leveraging data parallelisation, we could not train above a per-device batch size of 8 which meant that the number of requested GPUs had to be increased every couple of rounds in order to maintain manageable training and prediction times. The latter was particularly important since queuing times on the cluster could range anywhere from no waiting to waiting up to a week, potentially causing significant delays, particularly during the annotation process when the classifier had to be retrained weekly.

To increase the batch size and try further hyperparameter tuning, we made several attempts at parallelising the model using \citeposs{deepspeed} DeepSpeed optimisation library. The problem we encountered was that at the time of implementation, DeepSpeed appeared only to be configured for single user multi-GPU parallelisation which led to significant disruptions when implemented on a shared-user cluster. The behaviour we observed was that DeepSpeed reindexes available GPUs rather than inheriting the GPU numbers of the cluster, meaning that GPUs were called which were, in fact, not available, causing disruptions in other users' work. Despite trying a series of work around options in the form of intercepting the launch of DeepSpeed at an earlier point and avoiding this reassignment, we were unable to solve the problem from our end. We subsequently reported this issue to Microsoft but have not yet received a reply. 

Exploring options of how to achieve better performance using zero- and few-shot models \cite{brown:2020} as well as more powerful x-former models operating with more computational and memory efficiency \cite{tay:2020} is helpful. However, given our experience, it would also be desirable if future research shifted the focus more towards developing further options to parallelise models. Ideally, this would be done in a manner where software companies work together with research institutions in order to ensure better compatibility of available libraries and infrastructures. Another desirable direction of research would be for software companies to find a way to configure their cloud services in such a way that using them would not breach data protection requirements which do not allow uploading raw data to external parties.

A further infrastructural issue was encountered during deployment: at the moment, our hate speech classification API is running on university servers where we can offer the same infrastructure as was used during the configuration process and can thus provide an estimate of training and prediction times as well as costs. We are also still in charge of periodically checking classifier performance once the model is retrained. However, once our NGO partner and media companies decide to continue using the classifier without our input, they may no longer have the corresponding infrastructural set-up to run and maintain the classifier. Future work hoping to make contributions to both research and industry would thus benefit from implementing infrastructural frameworks which are tenable for industry partners. While research, like the work by \citet{tsankov:2022}, shows that this field is slowly gaining in importance, more needs to be done to facilitate easy deployment and long-term maintenance of AI systems.

\section{Ethics Statement}
\label{section:Ethics Statement}
While our dataset is more diverse than other hate speech benchmark sets, it is also prone to bias, a problem already addressed in numerous studies \cite{binns:2017, bender_friedman:2018, gebru:2018, holland:2018, davidson:2019, ranasinghe:2019, gorwa:2020, kim:2021} advocating for more transparency when it comes to constructing corpora from naturally occurring data. In our case, bias is introduced on four main levels: by the sources and platforms, and their respective representation in the dataset, by the manner in which training data were annotated throughout the annotation process, by the pre-trained model and by the prevalence of hate speech and distribution of target groups in the Swiss Hate Speech Corpus.

\subsection{Bias Stemming from Sources and Platforms}
Since our training set is mainly composed of data points from ON1 and Twitter, the classifier is well-tuned to examples stemming from these sources. When they occur during production, prediction is likely to achieve the performance results obtained during evaluation. However, as the cross-dataset and platform experiments show, the model does not yet generalise particularly well, which is partly conditioned on the fact that sources and platforms are represented differently in the dataset. ON2 and ON3, for instance, were only accessible to us towards the end of the study for final \textit{in situ} testing. Accordingly, a minority of the dataset covers comments from these two sources. Similarly, French tweets are -- compared to (Swiss) German tweets -- underrepresented in the corpus. While transfer learning certainly ensures that the classifier still achieves high performance, a more balanced representation of the different sources could lead to more model robustness.

At the same time, more generalisibility could be achieved by including into the training set examples from other online newspaper outlets and media platforms, like Facebook, Reddit, Signal or WhatsApp. Access to comment (or post) data from these platforms varies but since our classifier is periodically being retrained, these sources could easily be added in the future. The same applies for languages: our classification pipeline is based on mBERT, so if, in the future, data points in Italian, Romansh or other languages were to become available, they could easily be integrated into the existing dataset and -- through transfer learning -- potentially and naturally help to de-bias the set and enhance the performance. This approach is especially promising given the fact that French, Italian and Romansh belong to the same language family \cite{deshpande:2022}.

\subsection{Annotation Bias}
Defining hate speech is difficult in itself, but recognising it in real-life examples is even more difficult because the phenomenon keeps shifting and is entirely based on the concept of language usage and, thus, the context in which it was posted. The latter is not always available or clear enough, so decisions need to be made according to one's own best judgement. Existing research suggests that these judgments often depend on a series of annotator characteristics, like rational, background, linguistic competence and understanding of (the level of) hate speech \cite{wiegandetal:2019, mathew:2020, ousidhoum:2020, huang:2020, asogwa:2021}. 

In our case, the majority of annotators were university students, most of them in political science, and aged 20-27. With similar educational backgrounds and age groups, it is likely that annotations in the training set reflect how they interpret hate speech, especially since the majority of the comments and tweets in our training set were annotated by only a single annotator; for even with the guidelines and support we provided to them, the underlying understanding of when language usage crosses into hate speech is ultimately theirs. Furthermore, as outlined in  \hyperref[section:Annotation Process]{Appendix A} in more detail, our annotators themselves reported difficulties when annotating comments and tweets, which will have further influenced their judgements.

Quality controls were carried out to ensure coherence between the different annotation sets, but due to limitations in resources only a limited number were annotated a second or even a third time. Common themes we noticed throughout our deliberation rounds as well as during quality control and second annotations were that our annotators were particularly struggling to separate hate speech from various types of humour, enraged personal opinions, and opinions and feelings which do not reflect their own. In terms of humour, irony \cite{gibbs:2021}, sarcasm \cite{teh:2018} and euphemisms \cite{ayunts:2021} were often the sources of concern given their provocative nature and the difficulties associated in identifying them in the written, sometimes context-less, medium, where neither paraverbal communication nor body language can be used as indicators. Therefore, a more in-depth analysis of the annotated dataset would need to be carried out in the future to determine the extent to which our annotated dataset is biased towards the opinions of our research assistants.

In the meantime, volunteers working with our NGO partner as well as moderators who are currently using the deployed classification pipeline will, as central actors in the pipeline and with time, progressively de-bias or at least shift the bias of the system. If, as we envisage it, the network of our users can be extended to include NGOs and moderators from the Italian and Romansh speaking parts of Switzerland, of neighbouring countries and possibly even the Netherlands, Luxembourg and Belgium given the linguistic similarities between the respective languages and the issues they face with language- and context insensitive classification systems, even more diversification and further de-biasing of the model might be possible. 

\subsection{Model Bias}
While dataset bias has received most attention to-date, model bias is just as important an issue in configuring a hate speech detection system. As described in the \hyperref[section:Results and Discussion]{results section}, the pretrained models themselves carry various types of biases. We specifically encountered linguistic bias, likely introduced by the datasets on which the models were originally trained. In addition to fine-tuning the pretrained models, future research would benefit from placing more emphasis on de-biasing the models before fine-tuning them by following examples set by e.g. \citet{bolukbasi:2016, basta:2019}, who have attempted to remove gender bias from word embeddings. 

\subsection{Bias Stemming from the Prevalence of Hate Speech and Target Groups}
Part of the reason we decided to build a binary rather than a multilabel classifier was that the Swiss Hate Speech Corpus, as detailed in \hyperref[section:Annotation Process]{Appendix A}, includes a huge disparity in numbers of samples per target group category. Since these are naturally occurring data points, they reflect the prevalence of hate speech in Switzerland and its corresponding distribution across different target groups, but they also pose significant issues in terms of data imbalance for training multilabel classifiers. Attempts at balancing out the dataset with the help of a series of common regularisation techniques \cite{srivastava:2014, kukacka:2017, venturott:2020, feng:2021, venturott:2021} did not yield any useful results. Neither weighting classes, nor interpolating with SMOTE \cite{chawla:2011, fernandes:2018} worked.

Automatically augmenting the minority class with techniques like word splitting, syntax permutations, synonym replacement, insertions, deletions and substitutions as well as word2vec and contextual word embeddings with BERT, in order to upsample the minority class \cite{marivate:2019, dsa:2021} proved that label preservation was not possible, with mutations of many hate speech samples yielding adversarial non-hate speech instances under the hate speech label. It also showed that most instances of augmented data preserved the form, but not the often socially attributed meaning; that in the case of Swiss German dialects not even the form could be preserved; that, as \citet{shorten:2021} have already remarked, the techniques work better on longer than on shorter input sequences, leading to many problematic statements in short comments and especially in tweets; and that automatic spell checking was not configured for such a large amount of spelling errors and could, in some cases, not even identify the language of a specific comment or tweet. Given the sensitive nature of the subject and studies like \citet{longpre:2020}, who have shown that data augmentation techniques do not improve the performance of pretrained models capable of processing syntax and semantics, we did not employ these augmentation techniques.

Instead, more advanced data balancing techniques as described in e.g. \citet{mrksic:2016, shorten:2021, feng:2021, bayer:2021} and allowing for label preservation \cite{alzantot:2018, kobayashi:2018, rizos:2019} should be explored in the future. One of these methods, automatic back/round trip translation \cite{beddiar:2021}, has already been implemented on a subset of German tweets in a course paper by one of our annotators, yielding promising first results. If these techniques are successful or once more per-target group samples become available from our deployed model, a multilabel classifier, be this as a single or ensemble model, could be envisaged which would not bias the model towards certain target groups while neglecting others.

\section*{Acknowledgements}
\label{section:Acknowledgemnts}
We are grateful to the team at alliance F, whose StopHateSpeech campaign inspired this project; to InnoSuisse (Grant 46165.1 IP-SBM) and the Swiss Federal Office of Communications for funding; to the Swiss media companies who granted us access to their data; to Morgane Bonvallat, Laura Bronner, Philip Grech, Maël Kubli and Devin Routh for making valuable contributions to this study; and to our annotators Osama Abdullah, Maxime Bataillard, Cyril Bosshard, Florian Curvaia, Florian Eblenkamp, Marc Eggenberger, Stephanie Fierz, Selina Flöcklmüller, Rachel Kunstmann, Felicia Mändli, Anna Meisser, Mattia Mettler, Paula Moser, Alexandra Nagel, Dylan Paltra, Jonathan Progin, Jennifer Roberts, Viviane Vogel and Bruno Wüest for working tirelessly to help us produce our annotated Swiss Hate Speech Corpus.

\bibliography{references}

\begin{thebibliography}{150}
\expandafter\ifx\csname natexlab\endcsname\relax\def\natexlab#1{#1}\fi

\bibitem[{Abro et~al.(2020)Abro, Shaikh, Hussain~Khand, Ali, Khan, and
  Mujtaba}]{abro:2020}
Sindhu Abro, Sarang Shaikh, Zahid Hussain~Khand, Zafar Ali, Sajid Khan, and
  Ghulam Mujtaba. 2020.
\newblock \href {https://doi.org/10.14569/IJACSA.2020.0110861} {Automatic hate
  speech detection using machine learning: A comparative study}.
\newblock \emph{International Journal of Advanced Computer Science and
  Applications}, 11(August).

\bibitem[{Akiba et~al.(2019)Akiba, Sano, Yanase, Ohta, and Koyama}]{akiba:2019}
Takuya Akiba, Shotaro Sano, Toshihiko Yanase, Takeru Ohta, and Masanori Koyama.
  2019.
\newblock \href {https://doi.org/10.1145/3292500.3330701} {Optuna: A
  next-generation hyperparameter optimization framework}.
\newblock In \emph{Proceedings of the 25th ACM SIGKDD International Conference
  on Knowledge Discovery \& Data Mining}, KDD '19, page 2623–2631, New York,
  NY, USA. Association for Computing Machinery.

\bibitem[{Aljero and Dimililer(2021)}]{aljero:2021}
Mona Khalifa~A. Aljero and Nazife Dimililer. 2021.
\newblock \href {https://doi.org/10.3390/app112411684} {A novel stacked
  ensemble for hate speech recognition}.
\newblock \emph{Applied Sciences}, 11(24).

\bibitem[{Alsafari and Sadaoui(2021)}]{alsafari:2021}
Safa Alsafari and Samira Sadaoui. 2021.
\newblock \href {https://doi.org/10.1080/08839514.2021.1988443}
  {Semi-supervised self-training of hate and offensive speech from social
  media}.
\newblock \emph{Applied Artificial Intelligence}, 0(0):1--25.

\bibitem[{Alshaalan and Al-Khalifa(2020)}]{alshaalan-al-khalifa:2020}
Raghad Alshaalan and Hend Al-Khalifa. 2020.
\newblock \href {https://aclanthology.org/2020.wanlp-1.2} {Hate speech
  detection in saudi twittersphere: A deep learning approach}.
\newblock In \emph{Proceedings of the Fifth Arabic Natural Language Processing
  Workshop}, pages 12--23, Barcelona, Spain (Online). Association for
  Computational Linguistics.

\bibitem[{Aluru et~al.(2020)Aluru, Mathew, Saha, and Mukherjee}]{aluru:2020}
Sai~Saketh Aluru, Binny Mathew, Punyajoy Saha, and Animesh Mukherjee. 2020.
\newblock \href {http://arxiv.org/abs/2004.06465} {Deep learning models for
  multilingual hate speech detection}.
\newblock \emph{CoRR}, abs/2004.06465.

\bibitem[{Alzantot et~al.(2018)Alzantot, Sharma, Elgohary, Ho, Srivastava, and
  Chang}]{alzantot:2018}
Moustafa Alzantot, Yash Sharma, Ahmed Elgohary, Bo{-}Jhang Ho, Mani~B.
  Srivastava, and Kai{-}Wei Chang. 2018.
\newblock \href {http://arxiv.org/abs/1804.07998} {Generating natural language
  adversarial examples}.
\newblock \emph{CoRR}, abs/1804.07998.

\bibitem[{Arango et~al.(2019)Arango, P\'{e}rez, and Poblete}]{arango:2019}
Aym\'{e} Arango, Jorge P\'{e}rez, and Barbara Poblete. 2019.
\newblock \href {https://doi.org/10.1145/3331184.3331262} {Hate speech
  detection is not as easy as you may think: A closer look at model
  validation}.
\newblock In \emph{Proceedings of the 42nd International ACM SIGIR Conference
  on Research and Development in Information Retrieval}, SIGIR'19, page
  45–54, New York, NY, USA. Association for Computing Machinery.

\bibitem[{Asogwa and Onwuama(2021)}]{asogwa:2021}
N.U. Asogwa and M.E. Onwuama. 2021.
\newblock \href {https://doi.org/doi:10.1177/21582440211005772} {Hate speech
  and authentic personhood: Unveiling the truth}.
\newblock \emph{SAGE Open}.

\bibitem[{Ayunts and Paronyan(2021)}]{ayunts:2021}
A.~Ayunts and S.~Paronyan. 2021.
\newblock \href {https://doi.org/https://doi.org/10.7577/fleks.4173} {From
  euphemism to verbal aggression in british and armenian cultures: A
  cross-cultural pragmatic perspective}.
\newblock \emph{FLEKS - Scandinavian Journal of Intercultural Theory and
  Practice}, 7(1):26--42.

\bibitem[{Badjatiya et~al.(2017)Badjatiya, Gupta, Gupta, and
  Varma}]{badjatiya:2017}
Pinkesh Badjatiya, Shashank Gupta, Manish Gupta, and Vasudeva Varma. 2017.
\newblock \href {https://doi.org/10.1145/3041021.3054223} {Deep learning for
  hate speech detection in tweets}.
\newblock In \emph{Proceedings of the 26th International Conference on World
  Wide Web Companion}, WWW '17 Companion, page 759–760, Republic and Canton
  of Geneva, CHE. International World Wide Web Conferences Steering Committee.

\bibitem[{Barbieri et~al.(2021)Barbieri, Anke, and
  Camacho{-}Collados}]{barbieri:2021}
Francesco Barbieri, Luis~Espinosa Anke, and Jos{\'{e}} Camacho{-}Collados.
  2021.
\newblock \href {http://arxiv.org/abs/2104.12250} {{XLM-T:} {A} multilingual
  language model toolkit for twitter}.
\newblock \emph{CoRR}, abs/2104.12250.

\bibitem[{Barnidge et~al.(2019)Barnidge, Kim, Sherrill, Luknar, and
  Zhang}]{barnidge:2019}
Matthew Barnidge, Bumsoo Kim, Lindsey~A. Sherrill, Žiga Luknar, and Jiehua
  Zhang. 2019.
\newblock Perceived exposure to and avoidance of hate speech in various
  communication settings.
\newblock \emph{Telematics Informatics}, 44.

\bibitem[{Basile et~al.(2019)Basile, Bosco, Fersini, Nozza, Patti,
  Rangel~Pardo, Rosso, and Sanguinetti}]{basile:2019}
Valerio Basile, Cristina Bosco, Elisabetta Fersini, Debora Nozza, Viviana
  Patti, Francisco~Manuel Rangel~Pardo, Paolo Rosso, and Manuela Sanguinetti.
  2019.
\newblock \href {https://doi.org/10.18653/v1/S19-2007} {{S}em{E}val-2019 task
  5: Multilingual detection of hate speech against immigrants and women in
  {T}witter}.
\newblock In \emph{Proceedings of the 13th International Workshop on Semantic
  Evaluation}, pages 54--63, Minneapolis, Minnesota, USA. Association for
  Computational Linguistics.

\bibitem[{Basta et~al.(2019)Basta, Costa{-}juss{\`{a}}, and Casas}]{basta:2019}
Christine Basta, Marta~Ruiz Costa{-}juss{\`{a}}, and Noe Casas. 2019.
\newblock \href {http://arxiv.org/abs/1904.08783} {Evaluating the underlying
  gender bias in contextualized word embeddings}.
\newblock \emph{CoRR}, abs/1904.08783.

\bibitem[{Bayer et~al.(2021)Bayer, Kaufhold, and Reuter}]{bayer:2021}
Markus Bayer, Marc{-}Andr{\'{e}} Kaufhold, and Christian Reuter. 2021.
\newblock \href {http://arxiv.org/abs/2107.03158} {A survey on data
  augmentation for text classification}.
\newblock \emph{CoRR}, abs/2107.03158.

\bibitem[{Beddiar et~al.(2021)Beddiar, Jahan, and Oussalah}]{beddiar:2021}
Djamila~Romaissa Beddiar, Md~Saroar Jahan, and Mourad Oussalah. 2021.
\newblock \href {http://arxiv.org/abs/2106.04681} {Data expansion using back
  translation and paraphrasing for hate speech detection}.
\newblock \emph{CoRR}, abs/2106.04681.

\bibitem[{Bender and Friedman(2018)}]{bender_friedman:2018}
Emily~M. Bender and Batya Friedman. 2018.
\newblock \href {https://doi.org/10.1162/tacl_a_00041} {Data statements for
  natural language processing: Toward mitigating system bias and enabling
  better science}.
\newblock \emph{Transactions of the Association for Computational Linguistics},
  6:587--604.

\bibitem[{Binder et~al.(2020)Binder, Ueberwasser, and Stark}]{binder:2020}
Larissa Binder, Simone Ueberwasser, and Elisabeth Stark. 2020.
\newblock \href {https://doi.org/10.30687/978-88-6969-478-3/003} {Gendered hate
  speech in swiss {WhatsApp} messages}.
\newblock In \emph{Language, Gender and Hate Speech A Multidisciplinary
  Approach}. Fondazione Universit{\`{a}} Ca' Foscari.

\bibitem[{Binns et~al.(2017)Binns, Veale, Van~Kleek, and Shadbolt}]{binns:2017}
Reuben Binns, Michael Veale, Max Van~Kleek, and Nigel Shadbolt. 2017.
\newblock \href {http://arxiv.org/abs/1707.01477} {{Like trainer, like bot?
  Inheritance of bias in algorithmic content moderation}}.
\newblock \emph{CoRR}, abs/1707.01477.

\bibitem[{Bolukbasi et~al.(2016)Bolukbasi, Chang, Zou, Saligrama, and
  Kalai}]{bolukbasi:2016}
Tolga Bolukbasi, Kai-Wei Chang, James~Y. Zou, Venkatesh Saligrama, and Adam~T.
  Kalai. 2016.
\newblock \href
  {https://proceedings.neurips.cc/paper/2016/file/a486cd07e4ac3d270571622f4f316ec5-Paper.pdf}
  {Man is to computer programmer as woman is to homemaker? debiasing word
  embeddings}.
\newblock In \emph{Advances in Neural Information Processing Systems},
  volume~29. Curran Associates, Inc.

\bibitem[{Bosco et~al.(2018)Bosco, Dell’Orletta, Poletto, Sanguinetti, and
  Tesconi}]{bosco:2018}
Cristina Bosco, Felice Dell’Orletta, Fabio Poletto, Manuela Sanguinetti, and
  Maurizio Tesconi. 2018.
\newblock \href {https://doi.org/https://doi.org/10.4000/books.aaccademia.4503}
  {Overview of the evalita 2018 hate speech detection task}.
\newblock In \emph{EVALITA Evaluation of NLP and Speech Tools for Italian:
  Proceedings of the Final Workshop 12-13 December 2018, Naples [online].},
  Torino. Accademia University Press.

\bibitem[{Brown et~al.(2020)Brown, Mann, Ryder, Subbiah, Kaplan, Dhariwal,
  Neelakantan, Shyam, Sastry, Askell, Agarwal, Herbert{-}Voss, Krueger,
  Henighan, Child, Ramesh, Ziegler, Wu, Winter, Hesse, Chen, Sigler, Litwin,
  Gray, Chess, Clark, Berner, McCandlish, Radford, Sutskever, and
  Amodei}]{brown:2020}
Tom~B. Brown, Benjamin Mann, Nick Ryder, Melanie Subbiah, Jared Kaplan,
  Prafulla Dhariwal, Arvind Neelakantan, Pranav Shyam, Girish Sastry, Amanda
  Askell, Sandhini Agarwal, Ariel Herbert{-}Voss, Gretchen Krueger, Tom
  Henighan, Rewon Child, Aditya Ramesh, Daniel~M. Ziegler, Jeffrey Wu, Clemens
  Winter, Christopher Hesse, Mark Chen, Eric Sigler, Mateusz Litwin, Scott
  Gray, Benjamin Chess, Jack Clark, Christopher Berner, Sam McCandlish, Alec
  Radford, Ilya Sutskever, and Dario Amodei. 2020.
\newblock \href {http://arxiv.org/abs/2005.14165} {Language models are few-shot
  learners}.
\newblock \emph{CoRR}, abs/2005.14165.

\bibitem[{Br{\"u}gger et~al.(2009)Br{\"u}gger, Lalive, and
  Zweim{\"u}ller}]{bruegger:1990}
Beatrix Br{\"u}gger, Rafael Lalive, and Josef Zweim{\"u}ller. 2009.
\newblock \href {https://docs.iza.org/dp4283.pdf} {Does culture affect
  unemployment? evidence from the r{\"o}stigraben}.
\newblock \emph{IZA Institute of Labor Economics}, 4283:1--44.

\bibitem[{Budd et~al.(2019)Budd, Robinson, and Kainz}]{budd:2019}
Samuel Budd, Emma~C. Robinson, and Bernhard Kainz. 2019.
\newblock \href {http://arxiv.org/abs/1910.02923} {A survey on active learning
  and human-in-the-loop deep learning for medical image analysis}.
\newblock \emph{CoRR}, abs/1910.02923.

\bibitem[{Burnap and Williams(2016)}]{burnap:2016}
Pete Burnap and Matthew~L. Williams. 2016.
\newblock \href {https://doi.org/10.1140/epjds/s13688-016-0072-6} {Us and them:
  identifying cyber hate on twitter across multiple protected characteristics}.
\newblock \emph{EPJ Data Science}, 5(1):11.

\bibitem[{Burnap and Leighton~Williams(2014)}]{burnap:2014}
Peter Burnap and Matthew Leighton~Williams. 2014.
\newblock Hate speech, machine classification and statistical modelling of
  information flows on twitter: interpretation and communication for policy
  decision making.
\newblock URL: \url{https://orca.cardiff.ac.uk/65227/1/IPP2014-Burnap.pdf},
  access date 22.02.2022.

\bibitem[{Chan et~al.(2020)Chan, Schweter, and M{\"o}ller}]{chan:2020}
Branden Chan, Stefan Schweter, and Timo M{\"o}ller. 2020.
\newblock \href {https://doi.org/10.18653/v1/2020.coling-main.598}
  {{G}erman{'}s next language model}.
\newblock In \emph{Proceedings of the 28th International Conference on
  Computational Linguistics}, pages 6788--6796, Barcelona, Spain (Online).
  International Committee on Computational Linguistics.

\bibitem[{Chaudhry and Lease(2020)}]{chaudhry:2020}
Prateek Chaudhry and Matthew Lease. 2020.
\newblock \href {http://arxiv.org/abs/2012.09090} {You are what you tweet:
  Profiling users by past tweets to improve hate speech detection}.
\newblock \emph{CoRR}, abs/2012.09090.

\bibitem[{Chawla et~al.(2011)Chawla, Bowyer, Hall, and
  Kegelmeyer}]{chawla:2011}
Nitesh~V. Chawla, Kevin~W. Bowyer, Lawrence~O. Hall, and W.~Philip Kegelmeyer.
  2011.
\newblock \href {http://arxiv.org/abs/1106.1813} {{SMOTE:} synthetic minority
  over-sampling technique}.
\newblock \emph{CoRR}, abs/1106.1813.

\bibitem[{Chiril et~al.(2019)Chiril, Benamara~Zitoune, Moriceau, Coulomb-Gully,
  and Kumar}]{chiril:2019}
Patricia Chiril, Farah Benamara~Zitoune, V{\'e}ronique Moriceau, Marl{\`e}ne
  Coulomb-Gully, and Abhishek Kumar. 2019.
\newblock \href {https://aclanthology.org/2019.jeptalnrecital-court.21}
  {Multilingual and multitarget hate speech detection in tweets}.
\newblock In \emph{Actes de la Conf{\'e}rence sur le Traitement Automatique des
  Langues Naturelles (TALN) PFIA 2019. Volume II : Articles courts}, pages
  351--360, Toulouse, France. ATALA.

\bibitem[{{\c{C}}{\"o}ltekin(2020)}]{coltekin:2020}
{\c{C}}a{\u{g}}r{\i} {\c{C}}{\"o}ltekin. 2020.
\newblock \href {https://aclanthology.org/2020.lrec-1.758} {A corpus of
  {T}urkish offensive language on social media}.
\newblock In \emph{Proceedings of the 12th Language Resources and Evaluation
  Conference}, pages 6174--6184, Marseille, France. European Language Resources
  Association.

\bibitem[{Conneau et~al.(2019)Conneau, Khandelwal, Goyal, Chaudhary, Wenzek,
  Guzm{\'{a}}n, Grave, Ott, Zettlemoyer, and Stoyanov}]{conneau:2019}
Alexis Conneau, Kartikay Khandelwal, Naman Goyal, Vishrav Chaudhary, Guillaume
  Wenzek, Francisco Guzm{\'{a}}n, Edouard Grave, Myle Ott, Luke Zettlemoyer,
  and Veselin Stoyanov. 2019.
\newblock \href {http://arxiv.org/abs/1911.02116} {Unsupervised cross-lingual
  representation learning at scale}.
\newblock \emph{CoRR}, abs/1911.02116.

\bibitem[{Corazza et~al.(2020)Corazza, Menini, Cabrio, Tonelli, and
  Villata}]{corazza:2020}
Michele Corazza, Stefano Menini, Elena Cabrio, Sara Tonelli, and Serena
  Villata. 2020.
\newblock \href {https://doi.org/10.1145/3377323} {{A Multilingual Evaluation
  for Online Hate Speech Detection}}.
\newblock \emph{{ACM Transactions on Internet Technology}}, 20(2):1--22.

\bibitem[{Davidson et~al.(2019)Davidson, Bhattacharya, and
  Weber}]{davidson:2019}
Thomas Davidson, Debasmita Bhattacharya, and Ingmar Weber. 2019.
\newblock \href {https://doi.org/10.18653/v1/W19-3504} {Racial bias in hate
  speech and abusive language detection datasets}.
\newblock In \emph{Proceedings of the Third Workshop on Abusive Language
  Online}, pages 25--35, Florence, Italy. Association for Computational
  Linguistics.

\bibitem[{Davidson et~al.(2017)Davidson, Warmsley, Macy, and
  Weber}]{davidson:2017}
Thomas Davidson, Dana Warmsley, Michael~W. Macy, and Ingmar Weber. 2017.
\newblock Automated hate speech detection and the problem of offensive
  language.
\newblock In \emph{ICWSM}.

\bibitem[{Del~Vigna et~al.(2017)Del~Vigna, Cimino, Dell'Orletta, Petrocchi, and
  Tesconi}]{delvigna:2017}
Fabio Del~Vigna, Andrea Cimino, Felice Dell'Orletta, Marinella Petrocchi, and
  Maurizio Tesconi. 2017.
\newblock Hate me, hate me not: Hate speech detection on facebook.
\newblock In \emph{Proceedings of the First Italian Conference on Cybersecurity
  (ITASEC17)}.

\bibitem[{Deshpande et~al.(2022)Deshpande, Farris, and Kumar}]{deshpande:2022}
Neha Deshpande, Nicholas Farris, and Vidhur Kumar. 2022.
\newblock \href {http://arxiv.org/abs/2201.11294} {Highly generalizable models
  for multilingual hate speech detection}.
\newblock \emph{CoRR}, abs/2201.11294.

\bibitem[{Devlin et~al.(2019)Devlin, Chang, Lee, and Toutanova}]{devlin:2019}
Jacob Devlin, Ming-Wei Chang, Kenton Lee, and Kristina Toutanova. 2019.
\newblock \href {https://doi.org/10.18653/v1/N19-1423} {{BERT}: Pre-training of
  deep bidirectional transformers for language understanding}.
\newblock In \emph{Proceedings of the 2019 Conference of the North {A}merican
  Chapter of the Association for Computational Linguistics: Human Language
  Technologies, Volume 1 (Long and Short Papers)}, pages 4171--4186,
  Minneapolis, Minnesota. Association for Computational Linguistics.

\bibitem[{Djuric et~al.(2015)Djuric, Zhou, Morris, Grbovic, Radosavljevic, and
  Bhamidipati}]{djuric:2015}
Nemanja Djuric, Jing Zhou, Robin Morris, Mihajlo Grbovic, Vladan Radosavljevic,
  and Narayan Bhamidipati. 2015.
\newblock \href {https://doi.org/10.1145/2740908.2742760} {Hate speech
  detection with comment embeddings}.
\newblock In \emph{Proceedings of the 24th International Conference on World
  Wide Web}, WWW '15 Companion, page 29–30, New York, NY, USA. Association
  for Computing Machinery.

\bibitem[{Duzha et~al.(2021)Duzha, Casadei, Tosi, and Celli}]{duzha:2021}
Armend Duzha, Cristiano Casadei, Michael Tosi, and Fabio Celli. 2021.
\newblock \href {http://arxiv.org/abs/2107.05357} {Hate versus politics:
  Detection of hate against policy makers in italian tweets}.
\newblock \emph{CoRR}, abs/2107.05357.

\bibitem[{D’Sa et~al.(2021)D’Sa, Illina, Fohr, Klakow, and
  Ruiter}]{dsa:2021}
Aashwin~Geet D’Sa, Irina Illina, Dominique Fohr, Dietrich Klakow, and Dana
  Ruiter. 2021.
\newblock \href {https://doi.org/10.1007/978-3-030-83527-9_12} {Exploring
  conditional language model based data augmentation approaches for hate speech
  classification}.
\newblock In K.~Ekštein, F.~Pártl, and M.~Konopík, editors, \emph{Text,
  Speech, and Dialogue. TSD 2021. Lecture Notes in Computer Science}, volume
  12848. Springer.

\bibitem[{Enarsson et~al.(2022)Enarsson, Enqvist, and
  Naarttij{\"a}rvi}]{enarsson:2022}
Therese Enarsson, Lena Enqvist, and Markus Naarttij{\"a}rvi. 2022.
\newblock \href {https://doi.org/10.1080/13600834.2021.1958860} {Approaching
  the human in the loop - legal perspectives on hybrid human/algorithmic
  decision-making in three contexts}.
\newblock \emph{Information \& Communications Technology Law}, 31:1:123--153.

\bibitem[{Fanton et~al.(2021)Fanton, Bonaldi, Tekiroglu, and
  Guerini}]{fanton:2021}
Margherita Fanton, Helena Bonaldi, Serra~Sinem Tekiroglu, and Marco Guerini.
  2021.
\newblock \href {http://arxiv.org/abs/2107.08720} {Human-in-the-loop for data
  collection: a multi-target counter narrative dataset to fight online hate
  speech}.
\newblock \emph{CoRR}, abs/2107.08720.

\bibitem[{FDFA(2020)}]{fdfa:2020}
Federal Department of Foreign~Affairs FDFA. 2020.
\newblock Four languages, four cultures -- one country.
\newblock URL:
  \url{https://www.eda.admin.ch/eda/en/fdfa/fdfa/aktuell/newsuebersicht/2020/09/mehrsprachigkeit-1.html},
  access date 14.06.2022.

\bibitem[{Feng et~al.(2021)Feng, Gangal, Wei, Chandar, Vosoughi, Mitamura, and
  Hovy}]{feng:2021}
Steven~Y. Feng, Varun Gangal, Jason Wei, Sarath Chandar, Soroush Vosoughi,
  Teruko Mitamura, and Eduard~H. Hovy. 2021.
\newblock \href {http://arxiv.org/abs/2105.03075} {A survey of data
  augmentation approaches for {NLP}}.
\newblock \emph{CoRR}, abs/2105.03075.

\bibitem[{Fern\'{a}ndez et~al.(2018)Fern\'{a}ndez, Garc\'{\i}a, Herrera, and
  Chawla}]{fernandes:2018}
Alberto Fern\'{a}ndez, Salvador Garc\'{\i}a, Francisco Herrera, and Nitesh~V.
  Chawla. 2018.
\newblock {SMOTE for Learning from Imbalanced Data: Progress and Challenges,
  Marking the 15-Year Anniversary}.
\newblock \emph{Journal of Artificial Intelligence Research}, 61(1):863–905.

\bibitem[{Fersini et~al.(2018)Fersini, Nozza, and Rosso}]{fersini:2018}
Elisabetta Fersini, Debora Nozza, and Paolo Rosso. 2018.
\newblock Overview of the evalita 2018 task on automatic misogyny
  identification (ami).
\newblock In \emph{EVALITA@CLiC-it}.

\bibitem[{Florio et~al.(2021)Florio, Basile, and Patti}]{florio:2021}
Komal Florio, Valerio Basile, and Viviana Patti. 2021.
\newblock Hate speech and topic shift in the covid-19 public discourse on
  social media in italy.
\newblock In \emph{CLiC-it}, volume 3033 of \emph{{CEUR} Workshop Proceedings}.
  CEUR-WS.org.

\bibitem[{Fortuna and Nunes(2018)}]{fortuna:2019}
Paula Fortuna and S\'{e}rgio Nunes. 2018.
\newblock \href {https://doi.org/10.1145/3232676} {A survey on automatic
  detection of hate speech in text}.
\newblock \emph{ACM Comput. Surv.}, 51(4).

\bibitem[{Gama et~al.(2014)Gama, \v{Z}liobaitundefined, Bifet, Pechenizkiy, and
  Bouchachia}]{gama:2014}
Jo\~{a}o Gama, Indrundefined \v{Z}liobaitundefined, Albert Bifet, Mykola
  Pechenizkiy, and Abdelhamid Bouchachia. 2014.
\newblock \href {https://doi.org/10.1145/2523813} {A survey on concept drift
  adaptation}.
\newblock \emph{ACM Comput. Surv.}, 46(4).

\bibitem[{Gao and Huang(2017)}]{gao-huang:2017}
Lei Gao and Ruihong Huang. 2017.
\newblock \href {https://doi.org/10.26615/978-954-452-049-6_036} {Detecting
  online hate speech using context aware models}.
\newblock In \emph{Proceedings of the International Conference Recent Advances
  in Natural Language Processing, {RANLP} 2017}, pages 260--266, Varna,
  Bulgaria. INCOMA Ltd.

\bibitem[{Gebru et~al.(2018)Gebru, Morgenstern, Vecchione, Wortman~Vaughan,
  Wallach, Daum{\'{e}}~III, and Crawford}]{gebru:2018}
Timnit Gebru, Jamie Morgenstern, Briana Vecchione, Jennifer Wortman~Vaughan,
  Hanna~M. Wallach, Hal Daum{\'{e}}~III, and Kate Crawford. 2018.
\newblock \href {http://arxiv.org/abs/1803.09010} {Datasheets for datasets}.
\newblock \emph{CoRR}, abs/1803.09010.

\bibitem[{Geet~D'Sa et~al.(2020)Geet~D'Sa, Illina, and Fohr}]{geetdsa:2020}
Ashwin Geet~D'Sa, Irina Illina, and Dominique Fohr. 2020.
\newblock \href {https://hal.archives-ouvertes.fr/hal-03101938}
  {{Classification of Hate Speech Using Deep Neural Networks}}.
\newblock \emph{{Revue d'Information Scientifique \& Technique }}, 25(01).

\bibitem[{Geschke et~al.(2019)Geschke, Kla{\ss}en, Quent, and
  Richter}]{geschke:2019}
Daniel Geschke, Anja Kla{\ss}en, Matthias Quent, and Christoph Richter. 2019.
\newblock {HASS IM NETZ: Der schleichende Angriff auf unsere Demokratie. Eine
  bundesweite repr{\"a}sentative Untersuchung}.
\newblock \emph{Verf{\"u}gbar am}, 25:2020.

\bibitem[{Gibbs~Jr.(2021)}]{gibbs:2021}
Raymond~W. Gibbs~Jr. 2021.
\newblock \href {https://doi.org/10.1080/10926488.2020.1855944} {{“Holy Cow,
  My Irony Detector Just Exploded!” Calling Out Irony During The Coronavirus
  Pandemic}}.
\newblock \emph{Metaphor and Symbol}, 36(1):45--60.

\bibitem[{Gorwa et~al.(2020)Gorwa, Binns, and Christian}]{gorwa:2020}
Robert Gorwa, Reuben Binns, and Katzenbach Christian. 2020.
\newblock \href {https://doi.org/10.1177/2053951719897945} {{Algorithmic
  content moderation: Technical and political challenges in the automation of
  platform governance}}.
\newblock \emph{{Big Data \& Society}}.

\bibitem[{Hega(2001)}]{hega:2001}
Gunther~M. Hega. 2001.
\newblock \href {https://doi.org/10.1080/03057920120053210} {Regional identity,
  language and education policy in switzerland}.
\newblock \emph{Compare: A Journal of Comparative and International Education},
  31(2):205--227.

\bibitem[{Holland et~al.(2018)Holland, Hosny, Newman, Joseph, and
  Chmielinski}]{holland:2018}
Sarah Holland, Ahmed Hosny, Sarah Newman, Joshua Joseph, and Kasia Chmielinski.
  2018.
\newblock \href {http://arxiv.org/abs/1805.03677} {The dataset nutrition label:
  {A} framework to drive higher data quality standards}.
\newblock \emph{CoRR}, abs/1805.03677.

\bibitem[{Hosseini et~al.(2017)Hosseini, Kannan, Zhang, and
  Poovendran}]{hosseini:2017}
Hossein Hosseini, Sreeram Kannan, Baosen Zhang, and Radha Poovendran. 2017.
\newblock \href {http://arxiv.org/abs/1702.08138} {Deceiving google's
  perspective {API} built for detecting toxic comments}.
\newblock \emph{CoRR}, abs/1702.08138.

\bibitem[{Huang et~al.(2020)Huang, Xing, Dernoncourt, and Paul}]{huang:2020}
Xiaolei Huang, Linzi Xing, Franck Dernoncourt, and Michael~J. Paul. 2020.
\newblock \href {http://arxiv.org/abs/2002.10361} {Multilingual twitter corpus
  and baselines for evaluating demographic bias in hate speech recognition}.
\newblock \emph{CoRR}, abs/2002.10361.

\bibitem[{Jaki and De~Smedt(2019)}]{jaki-desmedt:2019}
Sylvia Jaki and Tom De~Smedt. 2019.
\newblock Right-wing german hate speech on twitter: Analysis and automatic
  detection.
\newblock \emph{ArXiv}, abs/1910.07518.

\bibitem[{Jauhiainen et~al.(2018)Jauhiainen, Jauhiainen, and
  Lind{\'e}n}]{jauhiainen:2018}
Tommi Jauhiainen, Heidi Jauhiainen, and Krister Lind{\'e}n. 2018.
\newblock \href {https://aclanthology.org/W18-3929} {{H}e{LI}-based experiments
  in {S}wiss {G}erman dialect identification}.
\newblock In \emph{Proceedings of the Fifth Workshop on {NLP} for Similar
  Languages, Varieties and Dialects ({V}ar{D}ial 2018)}, pages 254--262, Santa
  Fe, New Mexico, USA. Association for Computational Linguistics.

\bibitem[{Karan and {\v{S}}najder(2018)}]{karan-snajder:2018}
Mladen Karan and Jan {\v{S}}najder. 2018.
\newblock \href {https://doi.org/10.18653/v1/W18-5117} {Cross-domain detection
  of abusive language online}.
\newblock In \emph{Proceedings of the 2nd Workshop on Abusive Language Online
  ({ALW}2)}, pages 132--137, Brussels, Belgium. Association for Computational
  Linguistics.

\bibitem[{Kiela et~al.(2021)Kiela, Bartolo, Nie, Kaushik, Geiger, Wu, Vidgen,
  Prasad, Singh, Ringshia, Ma, Thrush, Riedel, Waseem, Stenetorp, Jia, Bansal,
  Potts, and Williams}]{kiela:2021}
Douwe Kiela, Max Bartolo, Yixin Nie, Divyansh Kaushik, Atticus Geiger,
  Zhengxuan Wu, Bertie Vidgen, Grusha Prasad, Amanpreet Singh, Pratik Ringshia,
  Zhiyi Ma, Tristan Thrush, Sebastian Riedel, Zeerak Waseem, Pontus Stenetorp,
  Robin Jia, Mohit Bansal, Christopher Potts, and Adina Williams. 2021.
\newblock \href {http://arxiv.org/abs/2104.14337} {Dynabench: Rethinking
  benchmarking in {NLP}}.
\newblock \emph{CoRR}, abs/2104.14337.

\bibitem[{Kim(2021)}]{kim:2021}
Jae~Yeon Kim. 2021.
\newblock \href {https://doi.org/10.31219/osf.io/chvgp} {{Power, Hate Speech,
  Machine Learning, and Intersectional Approach}}.
\newblock SocArXiv chvgp, Center for Open Science.

\bibitem[{Kobayashi(2018)}]{kobayashi:2018}
Sosuke Kobayashi. 2018.
\newblock \href {https://doi.org/10.18653/v1/N18-2072} {Contextual
  augmentation: Data augmentation by words with paradigmatic relations}.
\newblock In \emph{Proceedings of the 2018 Conference of the North {A}merican
  Chapter of the Association for Computational Linguistics: Human Language
  Technologies, Volume 2 (Short Papers)}, pages 452--457, New Orleans,
  Louisiana. Association for Computational Linguistics.

\bibitem[{Kokatnoor and Krishnan(2020)}]{kokatnoor:2020}
Sujatha~Arun Kokatnoor and Balachandran Krishnan. 2020.
\newblock \href {https://doi.org/10.1109/ICRCICN50933.2020.9296199} {Twitter
  hate speech detection using stacked weighted ensemble (swe) model}.
\newblock In \emph{2020 Fifth International Conference on Research in
  Computational Intelligence and Communication Networks (ICRCICN)}, pages
  87--92.

\bibitem[{Kukacka et~al.(2017)Kukacka, Golkov, and Cremers}]{kukacka:2017}
Jan Kukacka, Vladimir Golkov, and Daniel Cremers. 2017.
\newblock \href {http://arxiv.org/abs/1710.10686} {Regularization for deep
  learning: {A} taxonomy}.
\newblock \emph{CoRR}, abs/1710.10686.

\bibitem[{Kupi et~al.(2021)Kupi, Bodnar, Schmidt, and Posada}]{kupi:2021}
Maximilian Kupi, Michael Bodnar, Nikolas Schmidt, and Carlos~Eduardo Posada.
  2021.
\newblock dictnn: A dictionary-enhanced cnn approach for classifying hate
  speech on twitter.
\newblock \emph{ArXiv}, abs/2103.08780.

\bibitem[{Kwok and Wang(2013)}]{kwok-wang:2013}
Irene Kwok and Yuzhou Wang. 2013.
\newblock Locate the hate: Detecting tweets against blacks.
\newblock In \emph{Proceedings of the Twenty-Seventh AAAI Conference on
  Artificial Intelligence}, AAAI'13, page 1621–1622. AAAI Press.

\bibitem[{Laaksonen et~al.(2020)Laaksonen, Haapoja, Kinnunen, Nelimarkka, and
  P{\"o}yht{\"a}ri}]{laaksonen:2020}
Salla-Maaria Laaksonen, Jesse Haapoja, Teemu Kinnunen, Matti Nelimarkka, and
  Reeta P{\"o}yht{\"a}ri. 2020.
\newblock \href {https://doi.org/10.3389/fdata.2020.00003} {The datafication of
  hate: Expectations and challenges in automated hate speech monitoring}.
\newblock \emph{Frontiers in Big Data}, 3(3):1--16.

\bibitem[{Le et~al.(2019)Le, Vial, Frej, Segonne, Coavoux, Lecouteux, Allauzen,
  Crabb{\'{e}}, Besacier, and Schwab}]{le:2020}
Hang Le, Lo{\"{\i}}c Vial, Jibril Frej, Vincent Segonne, Maximin Coavoux,
  Benjamin Lecouteux, Alexandre Allauzen, Beno{\^{\i}}t Crabb{\'{e}}, Laurent
  Besacier, and Didier Schwab. 2019.
\newblock \href {http://arxiv.org/abs/1912.05372} {Flaubert: Unsupervised
  language model pre-training for french}.
\newblock \emph{CoRR}, abs/1912.05372.

\bibitem[{Leonardelli et~al.(2021)Leonardelli, Menini, Aprosio, Guerini, and
  Tonelli}]{leonardelli:2021}
Elisa Leonardelli, Stefano Menini, Alessio~Palmero Aprosio, Marco Guerini, and
  Sara Tonelli. 2021.
\newblock \href {http://arxiv.org/abs/2109.13563} {Agreeing to disagree:
  Annotating offensive language datasets with annotators' disagreement}.
\newblock \emph{CoRR}, abs/2109.13563.

\bibitem[{Longpre et~al.(2020)Longpre, Wang, and DuBois}]{longpre:2020}
Shayne Longpre, Yu~Wang, and Christopher DuBois. 2020.
\newblock \href {http://arxiv.org/abs/2010.01764} {How effective is
  task-agnostic data augmentation for pretrained transformers?}
\newblock \emph{CoRR}, abs/2010.01764.

\bibitem[{MacAvaney et~al.(2019)MacAvaney, Yao, Yang, Russell, Goharian, and
  Frieder}]{macavaney:2019}
Sean MacAvaney, Hao-Ren Yao, Eugene Yang, Katina Russell, Nazli Goharian, and
  Ophir Frieder. 2019.
\newblock \href {https://doi.org/10.1371/journal.pone.0221152} {Hate speech
  detection: Challenges and solutions}.
\newblock \emph{PloS one}, 14:e0221152.

\bibitem[{Majumder(2021)}]{french_roberta}
Abhilash Majumder. 2021.
\newblock {French RoBERTa}.
\newblock URL:
  \url{https://huggingface.co/abhilash1910/french-roberta/blob/main/README.md},
  access date 02.03.2022.

\bibitem[{Malik et~al.(2022)Malik, Pang, and van~den Hengel}]{malik:2022}
Jitendra~Singh Malik, Guansong Pang, and Anton van~den Hengel. 2022.
\newblock \href {http://arxiv.org/abs/2202.09517} {Deep learning for hate
  speech detection: a comparative study}.
\newblock \emph{CoRR}, abs/2202.09517.

\bibitem[{Malmasi and Zampieri(2017)}]{malmasi:2017}
Shervin Malmasi and Marcos Zampieri. 2017.
\newblock \href {https://doi.org/10.18653/v1/W17-1220} {{G}erman dialect
  identification in interview transcriptions}.
\newblock In \emph{Proceedings of the Fourth Workshop on {NLP} for Similar
  Languages, Varieties and Dialects ({V}ar{D}ial)}, pages 164--169, Valencia,
  Spain. Association for Computational Linguistics.

\bibitem[{Malmasi and Zampieri(2018)}]{malmasi:2018}
Shervin Malmasi and Marcos Zampieri. 2018.
\newblock \href {http://arxiv.org/abs/1803.05495} {Challenges in discriminating
  profanity from hate speech}.
\newblock \emph{CoRR}, abs/1803.05495.

\bibitem[{Mandl et~al.(2019)Mandl, Modha, Majumder, Patel, Dave, Mandlia, and
  Patel}]{mandletal:2019}
Thomas Mandl, Sandip Modha, Prasenjit Majumder, Daksh Patel, Mohana Dave,
  Chintak Mandlia, and Aditya Patel. 2019.
\newblock \href {https://doi.org/10.1145/3368567.3368584} {Overview of the
  hasoc track at fire 2019: Hate speech and offensive content identification in
  indo-european languages}.
\newblock In \emph{Proceedings of the 11th Forum for Information Retrieval
  Evaluation}, FIRE '19, page 14–17, New York, NY, USA. Association for
  Computing Machinery.

\bibitem[{Mandl et~al.(2020)Mandl, Modha, Shahi, Jaiswal, Nandini, Patel,
  Majumder, and Sch{\"{a}}fer}]{mandletal_2020:2020}
Thomas Mandl, Sandip Modha, Gautam~Kishore Shahi, Amit~Kumar Jaiswal, Durgesh
  Nandini, Daksh Patel, Prasenjit Majumder, and Johannes Sch{\"{a}}fer. 2020.
\newblock \href {http://arxiv.org/abs/2108.05927} {Overview of the {HASOC}
  track at {FIRE} 2020: Hate speech and offensive content identification in
  indo-european languages}.
\newblock \emph{CoRR}, abs/2108.05927.

\bibitem[{Mandl et~al.(2021)Mandl, Modha, Shahi, Madhu, Satapara, Majumder,
  Sch{\"{a}}fer, Ranasinghe, Zampieri, Nandini, and
  Jaiswal}]{mandletal_2021:2021}
Thomas Mandl, Sandip Modha, Gautam~Kishore Shahi, Hiren Madhu, Shrey Satapara,
  Prasenjit Majumder, Johannes Sch{\"{a}}fer, Tharindu Ranasinghe, Marcos
  Zampieri, Durgesh Nandini, and Amit~Kumar Jaiswal. 2021.
\newblock \href {http://arxiv.org/abs/2112.09301} {Overview of the {HASOC}
  subtrack at {FIRE} 2021: Hate speech and offensive content identification in
  english and indo-aryan languages}.
\newblock \emph{CoRR}, abs/2112.09301.

\bibitem[{Marivate and Sefara(2019)}]{marivate:2019}
Vukosi Marivate and Tshephisho Sefara. 2019.
\newblock \href {http://arxiv.org/abs/1907.03752} {Improving short text
  classification through global augmentation methods}.
\newblock \emph{CoRR}, abs/1907.03752.

\bibitem[{Mathew et~al.(2020)Mathew, Saha, Yimam, Biemann, Goyal, and
  Mukherjee}]{mathew:2020}
Binny Mathew, Punyajoy Saha, Seid~Muhie Yimam, Chris Biemann, Pawan Goyal, and
  Animesh Mukherjee. 2020.
\newblock \href {http://arxiv.org/abs/2012.10289} {Hatexplain: {A} benchmark
  dataset for explainable hate speech detection}.
\newblock \emph{CoRR}, abs/2012.10289.

\bibitem[{Melton et~al.(2020)Melton, Bagavathi, and Krishnan}]{melton:2020}
Joshua Melton, Arunkumar Bagavathi, and Siddhartha Krishnan. 2020.
\newblock Del-hate: A deep learning tunable ensemble for hate speech detection.
\newblock \emph{2020 19th IEEE International Conference on Machine Learning and
  Applications (ICMLA)}, pages 1015--1022.

\bibitem[{Meta(2022)}]{fb-hs-def}
Meta. 2022.
\newblock Facebook hate speech definition.
\newblock URL:
  \url{https://transparency.fb.com/en-gb/policies/community-standards/hate-speech/},
  access date 23.02.2022.

\bibitem[{Microsoft(2022{\natexlab{a}})}]{deepspeed}
Microsoft. 2022{\natexlab{a}}.
\newblock Deepspeed.
\newblock URL: \url{https://github.com/microsoft/deepspeed}, access date
  23.05.2022.

\bibitem[{Microsoft(2022{\natexlab{b}})}]{microsoft-hs-def}
Microsoft. 2022{\natexlab{b}}.
\newblock Microsoft hate speech definition.
\newblock URL: \url{https://www.microsoft.com/en-us/concern/hatespeech}, access
  date 23.02.2022.

\bibitem[{Mohiyaddeen and Siddiqi(2021)}]{mohiyaddeen:2021}
Mr. Mohiyaddeen and Sifatullah Siddiqi. 2021.
\newblock \href {https://doi.org/http://dx.doi.org/10.2139/ssrn.3887383}
  {Automatic hate speech detection: a literature review}.
\newblock \emph{SSRN}.

\bibitem[{Mondal et~al.(2017)Mondal, Silva, and Benevenuto}]{mondal:2017}
Mainack Mondal, Leandro~Ara\'{u}jo Silva, and Fabr\'{\i}cio Benevenuto. 2017.
\newblock \href {https://doi.org/10.1145/3078714.3078723} {A measurement study
  of hate speech in social media}.
\newblock In \emph{Proceedings of the 28th ACM Conference on Hypertext and
  Social Media}, HT '17, page 85–94, New York, NY, USA. Association for
  Computing Machinery.

\bibitem[{Mozafari et~al.(2019)Mozafari, Farahbakhsh, and
  Crespi}]{mozafari:2019}
Marzieh Mozafari, Reza Farahbakhsh, and No{\"{e}}l Crespi. 2019.
\newblock \href {http://arxiv.org/abs/1910.12574} {A bert-based transfer
  learning approach for hate speech detection in online social media}.
\newblock \emph{CoRR}, abs/1910.12574.

\bibitem[{Mrksic et~al.(2016)Mrksic, {\'{O}}~S{\'{e}}aghdha, Thomson, Gasic,
  Rojas{-}Barahona, Su, Vandyke, Wen, and Young}]{mrksic:2016}
Nikola Mrksic, Diarmuid {\'{O}}~S{\'{e}}aghdha, Blaise Thomson, Milica Gasic,
  Lina~Maria Rojas{-}Barahona, Pei{-}Hao Su, David Vandyke, Tsung{-}Hsien Wen,
  and Steve~J. Young. 2016.
\newblock \href {http://arxiv.org/abs/1603.00892} {Counter-fitting word vectors
  to linguistic constraints}.
\newblock \emph{CoRR}, abs/1603.00892.

\bibitem[{Mueller and Dardanelli(2014)}]{mueller:2014}
Sean Mueller and Paolo Dardanelli. 2014.
\newblock \href {https://doi.org/https://doi.org/10.3917/ripc.214.0083}
  {Langue, culture politique et centralisation en suisse}.
\newblock \emph{Revue internationale de politique comparée}, 21(4):83--104.

\bibitem[{Mullah and Zainon(2021)}]{mullah:2021}
Nanlir~Sallau Mullah and Wan Mohd Nazmee~Wan Zainon. 2021.
\newblock \href {https://doi.org/10.1109/ACCESS.2021.3089515} {Advances in
  machine learning algorithms for hate speech detection in social media: A
  review}.
\newblock \emph{IEEE Access}, 9:88364--88376.

\bibitem[{Nobata et~al.(2016)Nobata, Tetreault, Thomas, Mehdad, and
  Chang}]{nobata:2016}
Chikashi Nobata, Joel Tetreault, Achint Thomas, Yashar Mehdad, and Yi~Chang.
  2016.
\newblock \href {https://doi.org/10.1145/2872427.2883062} {Abusive language
  detection in online user content}.
\newblock In \emph{Proceedings of the 25th International Conference on World
  Wide Web}, WWW '16, page 145–153, Republic and Canton of Geneva, CHE.
  International World Wide Web Conferences Steering Committee.

\bibitem[{Nozza et~al.(2020)Nozza, Bianchi, and Hovy}]{nozza:2020}
Debora Nozza, Federico Bianchi, and Dirk Hovy. 2020.
\newblock \href {http://arxiv.org/abs/2003.02912} {What the [mask]? {Making}
  sense of language-specific {BERT} models}.
\newblock \emph{CoRR}, abs/2003.02912.

\bibitem[{Ousidhoum et~al.(2019)Ousidhoum, Lin, Zhang, Song, and
  Yeung}]{ousidhoum:2019}
Nedjma Ousidhoum, Zizheng Lin, Hongming Zhang, Yangqiu Song, and Dit-Yan Yeung.
  2019.
\newblock \href {https://doi.org/10.18653/v1/D19-1474} {Multilingual and
  multi-aspect hate speech analysis}.
\newblock In \emph{Proceedings of the 2019 Conference on Empirical Methods in
  Natural Language Processing and the 9th International Joint Conference on
  Natural Language Processing (EMNLP-IJCNLP)}, pages 4675--4684, Hong Kong,
  China. Association for Computational Linguistics.

\bibitem[{Ousidhoum et~al.(2020)Ousidhoum, Song, and Yeung}]{ousidhoum:2020}
Nedjma Ousidhoum, Yangqiu Song, and Dit-Yan Yeung. 2020.
\newblock \href {https://doi.org/10.18653/v1/2020.emnlp-main.199} {Comparative
  evaluation of label-agnostic selection bias in multilingual hate speech
  datasets}.
\newblock In \emph{Proceedings of the 2020 Conference on Empirical Methods in
  Natural Language Processing (EMNLP)}, pages 2532--2542, Online. Association
  for Computational Linguistics.

\bibitem[{Palakodety et~al.(2020)Palakodety, KhudaBukhsh, and
  Carbonell}]{palakodety:2020}
Shriphani Palakodety, Ashiqur~R. KhudaBukhsh, and Jaime~G. Carbonell. 2020.
\newblock \href {https://doi.org/10.3233/FAIA200456} {The refugee experience
  online: Surfacing positivity amidst hate}.
\newblock In Giuseppe~De Giacomo, Alejandro Catala, Bistra Dilkina, Michela
  Milano, Senén Barro, Alberto Bugarín, and Jérôme Lang, editors,
  \emph{ECAI 2020: Frontiers in Artificial Intelligence and Applications},
  pages 2925--2926. IOS Press.

\bibitem[{Pamungkas et~al.(2018)Pamungkas, Cignarella, Basile, and
  Patti}]{pamungkas:2018}
Endang~Wahyu Pamungkas, Alessandra~Teresa Cignarella, Valerio Basile, and
  Viviana Patti. 2018.
\newblock {Automatic identification of misogyny in English and Italian Tweet at
  EVALITA 2018 with a multilingual hate lexicon}.
\newblock In \emph{EVALITA@CLiC-it}, volume 2263 of \emph{{CEUR} Workshop
  Proceedings}. CEUR-WS.org.

\bibitem[{Pelicon et~al.(2021)Pelicon, Shekhar, Martinc, {\v{S}}krlj, Purver,
  and Pollak}]{pelicon:2021}
Andra{\v{z}} Pelicon, Ravi Shekhar, Matej Martinc, Bla{\v{z}} {\v{S}}krlj,
  Matthew Purver, and Senja Pollak. 2021.
\newblock \href {https://aclanthology.org/2021.hackashop-1.5} {Zero-shot
  cross-lingual content filtering: Offensive language and hate speech
  detection}.
\newblock In \emph{Proceedings of the EACL Hackashop on News Media Content
  Analysis and Automated Report Generation}, pages 30--34, Online. Association
  for Computational Linguistics.

\bibitem[{Pereira-Kohatsu et~al.(2019)Pereira-Kohatsu, Quijano~S{\'a}nchez,
  Liberatore, and Camacho-Collados}]{pereira-kohatsu:2019}
Juan~Carlos Pereira-Kohatsu, Lara Quijano~S{\'a}nchez, Federico Liberatore, and
  Miguel Camacho-Collados. 2019.
\newblock Detecting and monitoring hate speech in twitter.
\newblock \emph{Sensors (Basel, Switzerland)}, 19.

\bibitem[{Pires et~al.(2019)Pires, Schlinger, and Garrette}]{pires:2019}
Telmo Pires, Eva Schlinger, and Dan Garrette. 2019.
\newblock \href {http://arxiv.org/abs/1906.01502} {How multilingual is
  multilingual bert?}
\newblock \emph{CoRR}, abs/1906.01502.

\bibitem[{Poletto et~al.(2021)Poletto, Basile, Sanguinetti, Bosco, and
  Patti}]{poletto:2021}
Fabio Poletto, Valerio Basile, Manuela Sanguinetti, Cristina Bosco, and Viviana
  Patti. 2021.
\newblock Resources and benchmark corpora for hate speech detection: a
  systematic review.
\newblock \emph{Language Resources and Evaluation}, 55:477--523.

\bibitem[{Ranasinghe et~al.(2019)Ranasinghe, Zampieri, and
  Hettiarachchi}]{ranasinghe:2019}
Tharindu Ranasinghe, Marcos Zampieri, and Hansi Hettiarachchi. 2019.
\newblock \href {http://ceur-ws.org/Vol-2517/T3-3.pdf} {Brums at hasoc 2019:
  Deep learning models for multilingual hate speech and offensive language
  identification}.
\newblock In \emph{Working Notes of FIRE 2019 - Forum for Information Retrieval
  Evaluation, Kolkata, India, December 12-15, 2019}, volume 2517 of \emph{CEUR
  Workshop Proceedings}, pages 199--207. CEUR-WS.org.

\bibitem[{Rani et~al.(2020)Rani, Suryawanshi, Goswami, Chakravarthi, Fransen,
  and McCrae}]{rani-2020}
Priya Rani, Shardul Suryawanshi, Koustava Goswami, Bharathi~Raja Chakravarthi,
  Theodorus Fransen, and John~Philip McCrae. 2020.
\newblock \href {https://aclanthology.org/2020.trac-1.7} {A comparative study
  of different state-of-the-art hate speech detection methods in
  {H}indi-{E}nglish code-mixed data}.
\newblock In \emph{Proceedings of the Second Workshop on Trolling, Aggression
  and Cyberbullying}, pages 42--48, Marseille, France. European Language
  Resources Association (ELRA).

\bibitem[{Ribeiro et~al.(2020)Ribeiro, Wu, Guestrin, and Singh}]{ribeiro:2020}
Marco~Tulio Ribeiro, Tongshuang Wu, Carlos Guestrin, and Sameer Singh. 2020.
\newblock \href {https://doi.org/10.18653/v1/2020.acl-main.442} {Beyond
  accuracy: Behavioral testing of {NLP} models with {C}heck{L}ist}.
\newblock In \emph{Proceedings of the 58th Annual Meeting of the Association
  for Computational Linguistics}, pages 4902--4912, Online. Association for
  Computational Linguistics.

\bibitem[{Risch et~al.(2021)Risch, Stoll, Wilms, and Wiegand}]{rischetal:2021}
Julian Risch, Anke Stoll, Lena Wilms, and Michael Wiegand. 2021.
\newblock \href {https://aclanthology.org/2021.germeval-1.1} {Overview of the
  {G}erm{E}val 2021 shared task on the identification of toxic, engaging, and
  fact-claiming comments}.
\newblock In \emph{Proceedings of the GermEval 2021 Shared Task on the
  Identification of Toxic, Engaging, and Fact-Claiming Comments}, pages 1--12,
  Duesseldorf, Germany. Association for Computational Linguistics.

\bibitem[{Rizos et~al.(2019)Rizos, Hemker, and Schuller}]{rizos:2019}
Georgios Rizos, Konstantin Hemker, and Bj\"{o}rn Schuller. 2019.
\newblock \href {https://doi.org/10.1145/3357384.3358040} {Augment to prevent:
  Short-text data augmentation in deep learning for hate-speech
  classification}.
\newblock In \emph{Proceedings of the 28th ACM International Conference on
  Information and Knowledge Management}, CIKM '19, page 991–1000, New York,
  NY, USA. Association for Computing Machinery.

\bibitem[{Ross et~al.(2016)Ross, Rist, Carbonell, Cabrera, Kurowsky, and
  Wojatzki}]{ross:2016}
Bj{\"{o}}rn Ross, Michael Rist, Guillermo Carbonell, Ben Cabrera, Nils
  Kurowsky, and Michael Wojatzki. 2016.
\newblock \href {https://arxiv.org/pdf/1701.08118.pdf} {{Measuring the
  Reliability of Hate Speech Annotations: The Case of the European Refugee
  Crisis}}.
\newblock In \emph{Proceedings of NLP4CMC III: 3rd Workshop on Natural Language
  Processing for Computer-Mediated Communication}, pages 6--9.

\bibitem[{R{\"o}ttger et~al.(2021)R{\"o}ttger, Vidgen, Nguyen, Waseem,
  Margetts, and Pierrehumbert}]{roettger:2021}
Paul R{\"o}ttger, Bertram Vidgen, Dong Nguyen, Zeerak Waseem, Helen~Z.
  Margetts, and Janet~B. Pierrehumbert. 2021.
\newblock Hatecheck: Functional tests for hate speech detection models.
\newblock In \emph{ACL/IJCNLP}.

\bibitem[{Saleem et~al.(2017)Saleem, Dillon, Benesch, and Ruths}]{saleem:2017}
Haji~Mohammad Saleem, Kelly~P. Dillon, Susan Benesch, and Derek Ruths. 2017.
\newblock \href {http://arxiv.org/abs/1709.10159} {A web of hate: Tackling
  hateful speech in online social spaces}.
\newblock \emph{CoRR}, abs/1709.10159.

\bibitem[{Salminen et~al.(2018)Salminen, Almerekhi, Milenkovi{\'c}, Jung, An,
  Kwak, and Jansen}]{salminenetal:2018}
Joni Salminen, Hind Almerekhi, Milica Milenkovi{\'c}, {Soon Gyo} Jung, Jisun
  An, Haewoon Kwak, and {Bernard J.} Jansen. 2018.
\newblock Anatomy of online hate: Developing a taxonomy and machine learning
  models for identifying and classifying hate in online news media.
\newblock In \emph{12th International AAAI Conference on Web and Social Media,
  ICWSM 2018}, 12th International AAAI Conference on Web and Social Media,
  ICWSM 2018, pages 330--339. AAAI press.
\newblock Publisher Copyright: Copyright {\textcopyright} 2018, Association for
  the Advancement of Artificial Intelligence (www.aaai.org). All rights
  reserved.; 12th International AAAI Conference on Web and Social Media, ICWSM
  2018 ; Conference date: 25-06-2018 Through 28-06-2018.

\bibitem[{Salminen et~al.(2020)Salminen, Hopf, Chowdhury, Jung, Almerekhi, and
  Jansen}]{salminen:2020}
Joni Salminen, Maximilian Hopf, S.~A. Chowdhury, Soon-Gyo Jung, Hind Almerekhi,
  and Bernard~Jim Jansen. 2020.
\newblock Developing an online hate classifier for multiple social media
  platforms.
\newblock \emph{Human-centric Computing and Information Sciences}, 10:1--34.

\bibitem[{Salminen et~al.(2021)Salminen, Linarez, Jung, and
  Jansen}]{salminen:2021}
Joni Salminen, Maria~Jose Linarez, Soon-gyo Jung, and Bernard~J. Jansen. 2021.
\newblock \href {https://doi.org/10.1109/BESC53957.2021.9635436} {Online hate
  detection systems: Challenges and action points for developers, data
  scientists, and researchers}.
\newblock In \emph{2021 8th International Conference on Behavioral and Social
  Computing (BESC)}, pages 1--7.

\bibitem[{Schmidt and Wiegand(2017)}]{schmidt-wiegand:2017}
Anna Schmidt and Michael Wiegand. 2017.
\newblock \href {https://doi.org/10.18653/v1/W17-1101} {A survey on hate speech
  detection using natural language processing}.
\newblock In \emph{Proceedings of the Fifth International Workshop on Natural
  Language Processing for Social Media}, pages 1--10, Valencia, Spain.
  Association for Computational Linguistics.

\bibitem[{Shorten et~al.(2021)Shorten, Khoshgoftaar, and Furht}]{shorten:2021}
Connor Shorten, Taghi Khoshgoftaar, and Borko Furht. 2021.
\newblock \href {https://doi.org/10.1186/s40537-021-00492-0} {Text data
  augmentation for deep learning}.
\newblock \emph{Journal of Big Data}, 8.

\bibitem[{Siegel(2020)}]{siegel:2020}
Alexandra~A. Siegel. 2020.
\newblock \emph{Online Hate Speech}, SSRC Anxieties of Democracy, page 56–88.
  Cambridge University Press.

\bibitem[{Sood et~al.(2012)Sood, Churchill, and Antin}]{sood:2012}
Sara~Owsley Sood, Elizabeth~F. Churchill, and Judd Antin. 2012.
\newblock \href {https://doi.org/10.1002/asi.21690} {Automatic identification
  of personal insults on social news sites}.
\newblock \emph{The Journal of the Association for Information Science and
  Technology}, 63(2):270–285.

\bibitem[{Soral et~al.(2018)Soral, Bilewicz, and Winiewski}]{soral:2018}
Wiktor Soral, Michał Bilewicz, and Mikołaj Winiewski. 2018.
\newblock \href {https://doi.org/https://doi.org/10.1002/ab.21737} {Exposure to
  hate speech increases prejudice through desensitization}.
\newblock \emph{Aggressive Behavior}, 44(2):136--146.

\bibitem[{Srivastava et~al.(2014)Srivastava, Hinton, Krizhevsky, Sutskever, and
  Salakhutdinov}]{srivastava:2014}
Nitish Srivastava, Geoffrey Hinton, Alex Krizhevsky, Ilya Sutskever, and Ruslan
  Salakhutdinov. 2014.
\newblock Dropout: A simple way to prevent neural networks from overfitting.
\newblock \emph{J. Mach. Learn. Res.}, 15(1):1929–1958.

\bibitem[{Stevenson(1990)}]{stevenson:1990}
Patrick Stevenson. 1990.
\newblock \href {https://doi.org/10.1080/01434632.1990.9994413} {Political
  culture and intergroup relations in plurilingual switzerland}.
\newblock \emph{Journal of Multilingual \& Multicultural Development},
  11(3):227--255.

\bibitem[{Struss et~al.(2019)Struss, Siegel, Ruppenhofer, Wiegand, and
  Klenner}]{struss:2019}
Julia~Maria Struss, Melanie Siegel, Josef Ruppenhofer, Michael Wiegand, and
  Manfred Klenner. 2019.
\newblock Overview of {GermEval} task 2, 2019 shared task on the identification
  of offensive language.
\newblock In \emph{KONVENS}.

\bibitem[{Taradhita and Putra(2021)}]{taradhita:2021}
Dewa Ayu~Nadia Taradhita and I~Ketut Gede~Darma Putra. 2021.
\newblock Hate speech classification in indonesian language tweets by using
  convolutional neural network.
\newblock \emph{Journal of ICT Research and Applications}, 14:225--239.

\bibitem[{Tay et~al.(2020)Tay, Dehghani, Bahri, and Metzler}]{tay:2020}
Yi~Tay, Mostafa Dehghani, Dara Bahri, and Donald Metzler. 2020.
\newblock \href {http://arxiv.org/abs/2009.06732} {Efficient transformers: {A}
  survey}.
\newblock \emph{CoRR}, abs/2009.06732.

\bibitem[{Teh et~al.(2018)Teh, Ooi, Chan, and Chuah}]{teh:2018}
Phoey~Lee Teh, Pei~Boon Ooi, Nee~Nee Chan, and Yee~Kang Chuah. 2018.
\newblock A comparative study of the effectiveness of sentiment tools and human
  coding in sarcasm detection.
\newblock \emph{Journal of Systems and Information Technology}, 20(3):358--374.

\bibitem[{Thelwall et~al.(2010)Thelwall, Buckley, Paltoglou, Cai, and
  Kappas}]{thelwall:2010}
Mike Thelwall, Kevan Buckley, Georgios Paltoglou, Di~Cai, and Arvid Kappas.
  2010.
\newblock Sentiment in short strength detection informal text.
\newblock \emph{Journal of the Association for Information Science and
  Technology}, 61:2544--2558.

\bibitem[{Tita and Zubiaga(2021)}]{tita:2021}
Teodor Tita and Arkaitz Zubiaga. 2021.
\newblock \href {http://arxiv.org/abs/2111.00981} {Cross-lingual hate speech
  detection using transformer models}.
\newblock \emph{CoRR}, abs/2111.00981.

\bibitem[{Tontodimamma et~al.(2021)Tontodimamma, Nissi, Sarra, and
  Fontanella}]{tontodimamma:2021}
Alice Tontodimamma, Eugenia Nissi, Annalina Sarra, and Lara Fontanella. 2021.
\newblock \href {https://doi.org/https://doi.org/10.1007/s11192-020-03737-6}
  {Thirty years of research into hate speech: topics of interest and their
  evolution}.
\newblock \emph{Scientometrics}, 126:157--179.

\bibitem[{Tsankov et~al.(2022)Tsankov, Bielik, Vechev, and
  Krause}]{tsankov:2022}
Petar Tsankov, Pavol Bielik, Martin Vechev, and Andreas Krause. 2022.
\newblock {LatticeFlow}.
\newblock URL: \url{https://latticeflow.ai}, access date 23.02.2022.

\bibitem[{Twitter(2022)}]{twitter-hs-def}
Twitter. 2022.
\newblock Twitter hateful conduct definition.
\newblock URL:
  \url{https://help.twitter.com/en/rules-and-policies/hateful-conduct-policy},
  access date 23.02.2022.

\bibitem[{{United Nations}(2019)}]{UN-HS-def}
{United Nations}. 2019.
\newblock {United Nations Strategy and Plan of Action on Hate Speech}.
\newblock URL:
  \url{https://www.un.org/en/genocideprevention/documents/advising-and-mobilizing/Action_plan_on_hate_speech_EN.pdf},
  access date 23.02.2022.

\bibitem[{Van~Hee et~al.(2015)Van~Hee, Lefever, Verhoeven, Mennes, Desmet,
  De~Pauw, Daelemans, and Hoste}]{vanheeetal:2015}
Cynthia Van~Hee, Els Lefever, Ben Verhoeven, Julie Mennes, Bart Desmet, Guy
  De~Pauw, Walter Daelemans, and Veronique Hoste. 2015.
\newblock \href {https://aclanthology.org/R15-1086} {Detection and fine-grained
  classification of cyberbullying events}.
\newblock In \emph{Proceedings of the International Conference Recent Advances
  in Natural Language Processing}, pages 672--680, Hissar, Bulgaria. INCOMA
  Ltd. Shoumen, BULGARIA.

\bibitem[{Vashistha and Zubiaga(2020)}]{vashistha:2020}
Neeraj Vashistha and Arkaitz Zubiaga. 2020.
\newblock \href {https://doi.org/10.3390/info12010005} {Online multilingual
  hate speech detection: Experimenting with hindi and english social media}.
\newblock \emph{Information}, 12(1):5.

\bibitem[{Vaswani et~al.(2017)Vaswani, Shazeer, Parmar, Uszkoreit, Jones,
  Gomez, Kaiser, and Polosukhin}]{vaswani:2017}
Ashish Vaswani, Noam Shazeer, Niki Parmar, Jakob Uszkoreit, Llion Jones,
  Aidan~N. Gomez, Lukasz Kaiser, and Illia Polosukhin. 2017.
\newblock \href {http://arxiv.org/abs/1706.03762} {Attention is all you need}.
\newblock \emph{CoRR}, abs/1706.03762.

\bibitem[{Vengattil and Culliford(2022)}]{fb_adapted_hs_def:2022}
Munsif Vengattil and Elizabeth Culliford. 2022.
\newblock \href
  {https://www.reuters.com/world/europe/exclusive-facebook-instagram-temporarily-allow-calls-violence-against-russians-2022-03-10/}
  {Facebook allows war posts urging violence against russian invaders}.
\newblock In \emph{Reuters}.

\bibitem[{Venturott and Ciarelli(2020)}]{venturott:2020}
L\'{\i}gia~Iunes Venturott and Patrick~Marques Ciarelli. 2020.
\newblock \href {https://doi.org/10.1145/3428658.3431760} {Data augmentation
  for improving hate speech detection on social networks}.
\newblock In \emph{Proceedings of the Brazilian Symposium on Multimedia and the
  Web}, WebMedia '20, page 249–252, New York, NY, USA. Association for
  Computing Machinery.

\bibitem[{Venturott and Ciarelli(2021)}]{venturott:2021}
Ligia~Iunes Venturott and Patrick~Marques Ciarelli. 2021.
\newblock \href {https://doi.org/10.1007/978-3-030-86230-5_61} {{Application of
  Data Augmentation Techniques for Hate Speech Detection with Deep Learning}}.
\newblock In G.~Marreiros, F.S. Melo, N.~Lau, H.~Lopes~Cardoso, and L.P. Reis,
  editors, \emph{Progress in Artificial Intelligence. EPIA 2021. Lecture Notes
  in Computer Science}, volume 12981. Springer.

\bibitem[{Vidgen et~al.(2019)Vidgen, Harris, Nguyen, Tromble, Hale, and
  Margetts}]{vidgen:2019}
Bertie Vidgen, Alex Harris, Dong Nguyen, Rebekah Tromble, Scott Hale, and Helen
  Margetts. 2019.
\newblock \href {https://doi.org/10.18653/v1/W19-3509} {Challenges and
  frontiers in abusive content detection}.
\newblock In \emph{Proceedings of the Third Workshop on Abusive Language
  Online}, pages 80--93, Florence, Italy. Association for Computational
  Linguistics.

\bibitem[{Waseem(2016)}]{waseem:2016}
Zeerak Waseem. 2016.
\newblock \href {https://doi.org/10.18653/v1/W16-5618} {Are you a racist or am
  {I} seeing things? {A}nnotator influence on hate speech detection on
  {T}witter}.
\newblock In \emph{Proceedings of the First Workshop on {NLP} and Computational
  Social Science}, pages 138--142, Austin, Texas. Association for Computational
  Linguistics.

\bibitem[{Waseem and Hovy(2016)}]{waseem-hovy:2016}
Zeerak Waseem and Dirk Hovy. 2016.
\newblock \href {https://doi.org/10.18653/v1/N16-2013} {Hateful symbols or
  hateful people? {P}redictive features for hate speech detection on
  {T}witter}.
\newblock In \emph{Proceedings of the {NAACL} Student Research Workshop}, pages
  88--93, San Diego, California. Association for Computational Linguistics.

\bibitem[{Wenjie and Arkaitz(2021)}]{yin:2021}
Yin Wenjie and Zubiaga Arkaitz. 2021.
\newblock \href {http://arxiv.org/abs/2102.08886} {Towards generalisable hate
  speech detection: a review on obstacles and solutions}.
\newblock URL: \url{https://arxiv.org/pdf/2102.08886.pdf}, access date
  15.02.2022.

\bibitem[{Wiegand et~al.(2019)Wiegand, Ruppenhofer, and
  Kleinbauer}]{wiegandetal:2019}
Michael Wiegand, Josef Ruppenhofer, and Thomas Kleinbauer. 2019.
\newblock \href {https://doi.org/10.18653/v1/N19-1060} {{D}etection of
  {A}busive {L}anguage: the {P}roblem of {B}iased {D}atasets}.
\newblock In \emph{Proceedings of the 2019 Conference of the North {A}merican
  Chapter of the Association for Computational Linguistics: Human Language
  Technologies, Volume 1 (Long and Short Papers)}, pages 602--608, Minneapolis,
  Minnesota. Association for Computational Linguistics.

\bibitem[{Wiegand and Siegel(2018)}]{wiegand-siegel:2018}
Michael Wiegand and Melanie Siegel. 2018.
\newblock Overview of the {GermEval} 2018 shared task on the identification of
  offensive language.
\newblock In \emph{Proceedings of GermEval 2018, 14th Conference on Natural
  Language Processing (KONVENS 2018)}, Vienna, Austria.

\bibitem[{Wolf et~al.(2019)Wolf, Debut, Sanh, Chaumond, Delangue, Moi, Cistac,
  Rault, Louf, Funtowicz, and Brew}]{wolf:2019}
Thomas Wolf, Lysandre Debut, Victor Sanh, Julien Chaumond, Clement Delangue,
  Anthony Moi, Pierric Cistac, Tim Rault, R{\'{e}}mi Louf, Morgan Funtowicz,
  and Jamie Brew. 2019.
\newblock \href {http://arxiv.org/abs/1910.03771} {Huggingface's transformers:
  State-of-the-art natural language processing}.
\newblock \emph{CoRR}, abs/1910.03771.

\bibitem[{Youtube(2022)}]{youtube-hs-def}
Youtube. 2022.
\newblock Youtube hate speech definition.
\newblock URL: \url{https://support.google.com/youtube/answer/2801939?hl=en},
  access date 23.02.2022.

\bibitem[{Zampieri et~al.(2019)Zampieri, Malmasi, Nakov, Rosenthal, Farra, and
  Kumar}]{zampieri:2019}
Marcos Zampieri, Shervin Malmasi, Preslav Nakov, Sara Rosenthal, Noura Farra,
  and Ritesh Kumar. 2019.
\newblock \href {https://doi.org/10.18653/v1/N19-1144} {Predicting the type and
  target of offensive posts in social media}.
\newblock In \emph{Proceedings of the 2019 Conference of the North {A}merican
  Chapter of the Association for Computational Linguistics: Human Language
  Technologies, Volume 1 (Long and Short Papers)}, pages 1415--1420,
  Minneapolis, Minnesota. Association for Computational Linguistics.

\bibitem[{Zhang et~al.(2018)Zhang, Robinson, and Tepper}]{zhang:2018}
Ziqi Zhang, David Robinson, and Jonathan~A. Tepper. 2018.
\newblock Detecting hate speech on twitter using a convolution-gru based deep
  neural network.
\newblock In \emph{ESWC}.

\bibitem[{Zimmerman et~al.(2018)Zimmerman, Kruschwitz, and
  Fox}]{zimmermann:2018}
Steven Zimmerman, Udo Kruschwitz, and Chris Fox. 2018.
\newblock {Improving Hate Speech Detection with Deep Learning Ensembles}.
\newblock In \emph{Proceedings of the Eleventh International Conference on
  Language Resources and Evaluation (LREC 2018)}, Miyazaki, Japan. European
  Language Resources Association (ELRA).

\end{thebibliography}
\bibliographystyle{acl_natbib}

\newpage
\appendix

\section{Data Collection and Annotation Process}
\label{section:Annotation Process}

\subsection{Data Collection}
Our data, originally unlabelled, stems from three different sources:

\begin{enumerate}
    \item comments posted under online newspaper articles collected by our NGO partner (NGO)
    \item online newspaper comments directly donated by three Swiss national online newspaper outlets (ON1, ON2 and ON3)
    \item tweets collected using the Twitter API of ``politically interested users'', i.e. accounts following at least five Swiss newspapers or politicians.
\end{enumerate}

The first two sources donated their data to us, while we collected tweets using the Twitter API. Particularly important for our study is source number two given that it includes published, moderated and deleted comments from three different media outlets. Each of the three ONs attracts different kinds of audiences in terms of age, and educational and sociopolitical backgrounds which influence the types of conversations, writing styles and uses of language more generally. Readers of and commentators in ON1 and ON2 display more similar characteristics than do those in ON3, as was also reflected in our post-deployment cross-dataset experiment results detailed in \hyperref[table:Table 4]{Table 4}. Using all three sources thus adds to the diversity of our dataset.

\subsection{Annotators}
Annotators for (Swiss) German were 16 Bachelor and Master students in political science aged 20 to 27 years, of whom 15 are native speakers of (Swiss) German; annotators for French examples were three students from various disciplines in the natural sciences, all native speakers of French and 20 to 25 years of age. Differences in mother tongues and directions of study introduce some diversity, but overall, our annotators form a homogeneous group.

\subsection{Annotator Instructions}
\begin{figure}
\centering
\includegraphics[width=0.5\textwidth]{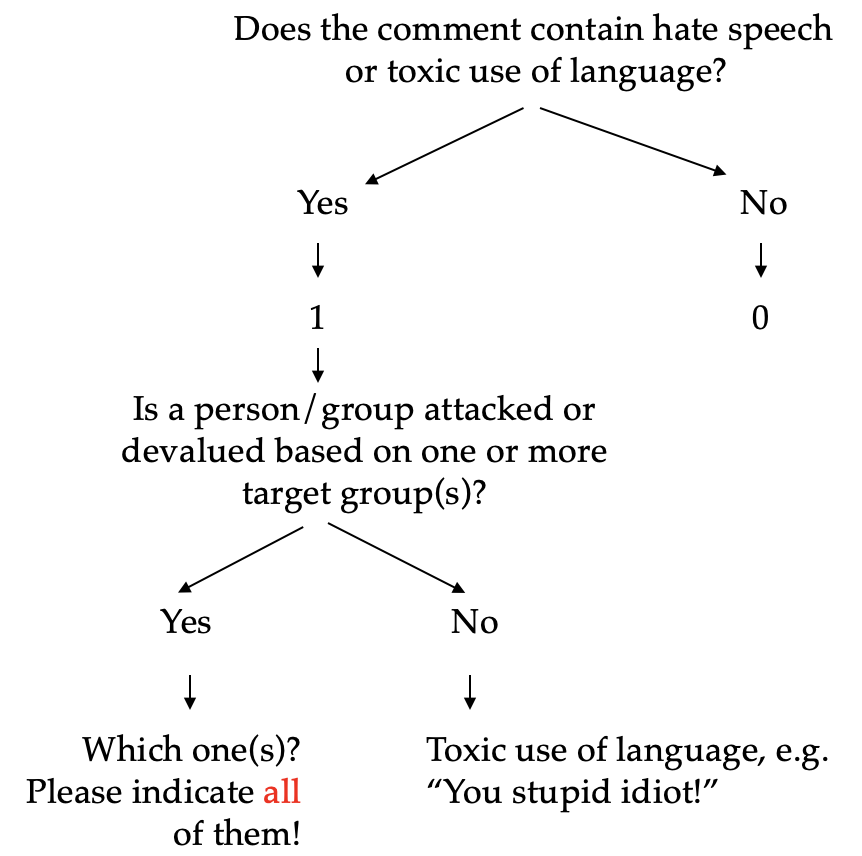}
\caption{Annotation Scheme}
\label{figure:Figure 1}
\end{figure}

All annotators attended a workshop introducing them to the annotation scheme depicted in \hyperref[figure:Figure 1]{Figure 1}. As can be seen, annotators were instructed first to decide whether or not a comment or tweet contains hate speech or toxic speech and to enter the corresponding number in the label column of their excel sheets. In case the comment or tweet contains hate speech, they were asked to indicate all groups against which the hate speech is targeted. Important to note in this scheme is that while a comment or tweet may be classified as targeting multiple groups, toxic speech is exclusive, meaning that once a comment or tweet is deemed to be toxic, it cannot be assigned to any other category. It is, however, still marked as 1 in the label column.

\subsection{Iterative Labelling}
The annotation process relied on iterative labelling as used by e.g. \citet{kiela:2021} and took full advantage of our human-in-the-loop hate speech classification pipeline with the exception that the human checks were performed by research assistants (RAs) and that retraining was carried out manually \cite[cf. also][]{fanton:2021}.

The annotation process took place in five waves over a period of 12 months. Before the first wave, a set 0 of collected (Swiss) German comments was annotated by the NGO community. The first two waves of annotations each lasted 9 weeks. During the first, (Swiss) German data was annotated, during the second, French data. Since the initial annotations revealed that our data features many non-hate speech and borderline cases, but hardly any clear hate speech cases, samples distributed for annotation included only those comments and tweets which were classified as having the highest hate speech probability. By seeing increasingly more instances of hate speech, the classifier gradually learnt to draw a clearer decision boundary between hate speech and non-hate speech.

At the beginning of the first two waves, a member of our team therefore annotated a sample of the classified comments with the highest hate speech probability. From then on, the two waves followed weekly cycles involving these steps:

\begin{enumerate}
    \item (Re)training the model using a class-balanced set drawn from all the data annotated up to that point.
    \item Making predictions on the (remaining) donated comments (first seven weeks) and tweets (final two weeks).
    \item From the positively classified comments and tweets, a subset is drawn in descending order of hate speech probability and is divided into smaller sets.
    \item The sets are distributed to RAs for annotation.
    \item While the sets are being annotated, a team member carries out quality control on the sets annotated in the previous week.
    \item The quality-controlled sets are added to the rest of the training set.
    \item The process begins anew until the final set is annotated and checked.
\end{enumerate}

For reasons of time and infrastructure, each comment and tweet was annotated by a single RA with the exception that in each weekly RA set, 60 identical comments or tweets were included for quality control. The third wave of annotations lasted for a week and was carried out on a set of (Swiss) German tweets. In this case, each tweet was annotated three times. 

The fourth wave of annotations took place post-deployment as part of a bigger experiment on counter speech strategies on Twitter and lasted for 15 weeks. Given this setup, single annotations were carried out, with a member of our team carrying out second annotations, this time not just of the tweets with highest hate speech probability, but taking -- for the first time -- advantage of the full prediction probability scale, i.e. from 1.0 to 0.5. The results of this manner of proceeding are summarised in \hyperref[table:Table 6]{Table 6} and highlight that there is a direct correlation between the number of hate speech tweets, their quality and the probability with which they were classified as hate speech: the further down the hate speech probability scale we go, the fewer actual hate speech tweets we find; and those we do find, are often contested, borderline or problematic cases.

The fifth wave of annotations also took place post-deployment, lasted for 9 weeks and involved three 3-week cycles through our human-in-the-loop hate speech classification pipeline. Each sample was annotated by two different RAs with a third RA making the final call on examples on which the first two could not agree. 

In cases where multiple annotations were available for a single example, unweighted linear averaging was applied using majority ruling. To avoid falling into the doctrinal paradox, the resulting final annotations were manually checked for consistency during quality control, either by RAs themselves in the third and fourth waves or by a member of our team in the other waves. 

The final result of this annotation process is, as shown in \hyperref[table:Table 12]{Table 12}, a large corpus of over 422k unique data points annotated for the binary label hate speech, the 10 multilabel target groups and the binary category toxic speech. This corpus is the first of its kind and is, when compared to other benchmark hate speech and toxic language corpora \cite{waseem-hovy:2016, davidson:2017, basile:2019}, one of -- if not the -- largest.

\subsection{Quality Control}
Quality control was carried out in terms of both quantity and quality. For the quantitative checks, we used Krippendorff's alpha to measure inter-rater reliability. Depending on the annotation wave, the results of this measurement oscillated between 0.3-0.4, which -- while fairly low -- is not surprising given the challenging and subjective nature of the task \cite{kwok-wang:2013, ross:2016, waseem:2016, delvigna:2017, malmasi:2018, mandletal:2019, ousidhoum:2019, rani-2020}. Consequently, bi-weekly deliberation rounds were introduced to discuss difficult cases. In waves where this was not possible, peer respect was used, with a member of our team carrying out second annotations on the positively annotated comments and tweets. While this meant mostly checking that the annotation scheme was followed as closely as possible and that all annotators classified the same type of hate speech in the same category, there were cases where a different annotation resulted after quality control. In addition to these quality checks, a subset of negatively classified comments with lowest non-hate speech probabilities was given out for annotation during the first wave (TN). The purpose of this set was to gain insight into the number of false negatives among the comments with lowest non-hate speech probability.

\subsection{Challenges and Heterogeneity of the Swiss Hate Speech Corpus}
Most challenges annotators experienced relate to cases which are so firmly embedded in the Swiss context that their negative connotations are only recognisable through subtexts. Examples of such expressions are listed in \hyperref[table:Table 7]{Table 7} and show that most of these instances target groups based on their nationality or political orientation. 

\begin{table*}[h]
\small
\noindent{\begin{center}\begin{tabular}{l l l l}
\hline
\textbf{Lang} & \textbf{Hate Speech Example} & \textbf{Translation}\\
\hline
\hline
G & \shortstack[l]{Bin i dr einzig, wo gseht us welere\\ Bevölkerigsgruppe dr chunt (\textit{Turnhose}). \\ Wiit muäs mr nied suächä, der \textit{geleaste}\\ \textit{BMW} ist nicht weit.} & \shortstack[l]{Am I the only one who sees which people\\ he belongs to (\textit{gym shorts}). One does not\\ have to look for long, the \textit{leased BMW}\\ can't be far.} \\
\\
& \shortstack[l]{Diese \textit{Balkonerin} soll sich in ihre heimat\\ \textit{Balkon} verschwindisieren.} & \shortstack[l]{That \textit{female balcony} should return\\ to her country \textit{balcony}.} \\
\\
& \shortstack[l]{\textit{grüne socken} sind wie \textit{wassermelonen}, aussen grün,\\ innen rot.} & \shortstack[l]{\textit{green socks} are like \textit{watermelons},\\ green on the outside, red on the inside.} \\
\\
\hline
\\
FR & \shortstack[l]{Il faut aussi savoir si tu es \textit{espingouin}\\ ou \textit{rital}?} & \shortstack[l]{You have to know whether you are\\\textit{Spanish} or \textit{Italian}?} \\
\\
& \shortstack[l]{Elle aura ce qu'elle mérite et peut-être\\ que ça lui apprendra à fermer sa grande \\g\#*"© de \textit{frouzes}!} & \shortstack[l]{She'll have what she deserves and\\ maybe that'll teach her to shut her\\ \textit{French} mouth!} \\
\\
& \shortstack[l]{de Genève les pleureuses \textit{shadoks} \\sont de retour ?} & \shortstack[l]{from Geneva, the winy \textit{French} are back?} \\
\\
& \shortstack[l]{Le \textit{seutch} tu peux le\\ garder.} & You can keep the \textit{French}. \\
\\
& Allez dire ça à ces foutus \textit{bourbines} & \shortstack[l]{Go tell this to those damn \textit{Swiss Germans}} \\
\hline
\end{tabular}\end{center}}\caption{Contextual Hate Speech}
\label{table:Table 7}
\end{table*}

Some of the contextual hate speech instances rely on the use of specific words (\textit{rital}, \textit{frouze}) or a distortion of words (\textit{Balkonerin}) to designate a particular target group, while others imply a target group through expressing a certain type of behaviour or other distinct characteristic. The notions \textit{(leased) BMW} and \textit{Turnhosen}, for instance, are derogatory terms for people, mostly men, from the Balkans on the basis that many of them wear gym shorts and drive BMW cars; and the expression \textit{grün-rote Wassermelone}, when used in a political context, attacks proponents of the green party by accusing them in a derogatory manner of pretending to be green, when in fact they are red, i.e. a socialist party.

In cases where RAs annotated comments, they were able to gauge the respective contexts more easily: comments sections are often not limited in the number of words or characters users can post, so individual comments are on average longer than tweets and provide more pieces of background information. RAs could also consult the title and text of the articles to gain additional context, which was not possible when annotating tweets. A further issue with tweets was that many of them were about events in Germany or Austria, using expressions and insults which are specific to these particular contexts and with which our RAs were not familiar.

While, with time and experience of these contexts, subtexts and jargon, RAs were able more easily to identify such difficult cases and the corresponding target groups, they also reported that through continual exposure to hateful and toxic statements, they felt that they were becoming increasingly desensitised to them, an observation already described by \citet{soral:2018} and \citet{barnidge:2019}. As a consequence, RAs noticed that there were cases which they would themselves annotate differently, once they re-read them at a later point in time.

RAs also reported that there remain many borderline cases, mostly involving implicit statements, terms which are potentially reclaimed by a minority group \cite{malmasi:2018}, strongly phrased personal opinions, sarcasm and irony, on which they could, even after discussing them, not agree. Another issue RAs faced concerns the diversity of languages and language varieties present in the dataset. Since hardly any guidelines are given on social media platforms as to the use of language, our dataset features a mix of the Swiss national languages, other languages commonly spoken in Switzerland (e.g. English and Portuguese) as well as the many dialects and varieties of Swiss German, whereby comments and tweets in German mostly use Standard German. Important to note is that both entire comments in a certain language or dialect variety and comments where several languages and dialect varieties are used simultaneously are present. 

While we were able to filter tweets according to language and could minimise this phenomenon, this was not always possible for comments below online newspaper articles. This is why, in addition to the linguistic difficulties, like different vocabulary, sentence structures and so on, which this change in languages and varieties poses, there are occasionally (parts of) comments which are written in a different script corresponding to the language being used (e.g. Arabic, Cyrillic or Greek scripts), sometimes triggering a different direction of reading (cf. Arabic). In addition to these language specific changes, the dataset also displays many graphic alterations deliberately introduced by users who -- presumably to game the system \cite{hosseini:2017} -- permute characters, words, sentence structures and make increased use of acronyms and abbreviations. This mix of codes and scripts was challenging for RAs who either decided not to annotate the comment or tweet in question or who spent a significant amount of time trying to research its content or translate it with the help of e.g. DeepL or Google Translate.

\subsection{The Swiss Hate Speech Corpus in Numbers}
The annotations generated in each wave of this process are summarised in \hyperref[table:Table 10]{Table 10}. As can be seen, of the 442782 annotations, 112337 are hate or toxic speech (HS\&Tox). Of these, 26815, i.e. approximately 24\%, were deemed toxic speech (Tox) and 85522 are hate speech (HS) against a total of 109812 groups (All) from 1 or more of the 10 possible (multilabel) target groups, suggesting that each targeted comment or tweet is, on average, targeting approximately 1.3 groups. Differences in target group distribution emerge when comparing the ONs and Twitter as well as the two languages. The most targeted groups in the ONs are nationality followed by sex, politics and social status, and on Twitter politics followed by other targets, nationality and social status. Twitter too features significantly higher numbers of toxic speech than do the ONs. 

Even though the groups which are most targeted are almost the same in both languages, the distribution of attacks among them varies, as shown in \hyperref[table:Table 13]{Table 13}: from the total of targeted hate speech comments in the German ONs, approximately 39\% feature hate speech against nationality, 32\% against politics, 14\% against other target groups, 13\% against sex and 12\% against social status compared to approximately 46\% nationality, 28\% sex, 20\% politics and 17\% social status in French. Similarly for Twitter: German has approximately 62\% hate speech against politics, 18\% against other targets and 14\% against nationality and social status, while French has 49\% against politics, 19\% against nationality and 18\% against sex and social status. These numbers suggest that French comments and tweets target politics to a lesser degree, and nationality, sex and social status more often than the Germans do. The same is true for impairments and appearance, while the reverse holds for the category `other' where more German ON comments can be found. Despite the fact that  the two languages feature similar numbers of attacks against age, gender and religion, these results confirm that language and sociocultural and political factors play a vital role in the frequency, distribution and nature of hate speech in Switzerland, and that varied distributions of them in the training set have the potential to bias the system.

\section{Results Tables}
\label{section:Tables}


\begin{table*}
\tiny
\begin{center}
\noindent{\begin{center}\begin{tabular}{l l l}
\hline
\hline
\textbf{Description} & \textbf{Details} & \textbf{Studies}\\
\hline
\hline
Input data & tweets & \citet{waseem-hovy:2016}\\
& & \citet{waseem:2016} \\
& & \citet{badjatiya:2017} \\
& & \citet{davidson:2017} \\
& & \citet{malmasi:2018} \\
& & \citet{pereira-kohatsu:2019} \\
& comments & \citet{sood:2012} \\
& & \citet{kwok-wang:2013} \\
& & \citet{saleem:2017} \\
& & \citet{delvigna:2017} \\
& & \citet{rani-2020} \\
\hline
Preprocessing & automatic translation into English & \citet{aluru:2020} \\
\hline
Languages & monolingual (mostly English) & \citet{malmasi:2018} \\
& & \citet{rani-2020} \\
& & \citet{aljero:2021} \\
& monolingual non-English & \citet{delvigna:2017} \\
& & \citet{pereira-kohatsu:2019} \\
& & \citet{coltekin:2020} \\
& & \citet{alshaalan-al-khalifa:2020} \\
& & \citet{taradhita:2021} \\
& multilingual & \citet{chiril:2019} \\
& & \citet{ousidhoum:2019} \\
& & \citet{aluru:2020} \\
& & \citet{rani-2020} \\
& & \citet{vashistha:2020} \\
& & \citet{nozza:2020} \\
& & \citet{tita:2021} \\
\hline
Classifier types & binary & \citet{malmasi:2018} \\
& & \citet{rani-2020} \\
& & \citet{aljero:2021} \\
& multilabel/-class & \citet{waseem-hovy:2016} \\
& & \citet{ross:2016} \\
& & \citet{davidson:2017} \\
& & \citet{salminenetal:2018} \\
\hline
Model architectures & statistical & \citet{saleem:2017} \\
& & \citet{pereira-kohatsu:2019} \\
& deep learning and transformers & \citet{badjatiya:2017} \\
& & \citet{zhang:2018} \\
& & \citet{zampieri:2019} \\
& & \citet{mozafari:2019} \\
& & \citet{geetdsa:2020} \\
& & \citet{kupi:2021} \\
& & \citet{rani-2020} \\
& & \citet{malik:2022} \\
& single and ensemble models & \citet{malmasi:2018} \\
& & \citet{kokatnoor:2020} \\
& & \citet{gao-huang:2017} \\
& & \citet{zimmermann:2018} \\
& & \citet{melton:2020} \\
& & \citet{aljero:2021} \\
& sentiment analysis & \citet{thelwall:2010} \\
& & \citet{sood:2012} \\
& & \citet{vanheeetal:2015} \\
\hline
Feature engineering
& surface-level features & \citet{mondal:2017} \\
& word embeddings & \citet{djuric:2015} \\
& & \citet{nobata:2016} \\
& LDA & \citet{saleem:2017} \\
& additional features to text & \citet{davidson:2017} \\
& & \citet{gao-huang:2017} \\
& & \citet{chaudhry:2020} \\
\hline
Categories & hate speech, toxic, profanity & \citet{davidson:2017} \\
& & \citet{malmasi:2018} \\
& specific target groups & \citet{kwok-wang:2013} \\
& & \citet{burnap:2014} \\
& & \citet{burnap:2016} \\
& & \citet{waseem-hovy:2016} \\
& & \citet{fersini:2018} \\
& & \citet{pamungkas:2018}\\
& & \citet{jaki-desmedt:2019} \\
& & \citet{chiril:2019} \\
& & \citet{basile:2019} \\
& & \citet{duzha:2021} \\
& \textit{voice-for-the-voiceless} & \citet{palakodety:2020} \\
\hline
\end{tabular}\end{center}}
\end{center}
\caption{Related Work}
\label{table:Table 8}
\end{table*}




\begin{table*}
\tiny
\begin{center}
\noindent{\begin{center}\begin{tabular}{l l r l r l l l}
\hline
\hline
{\bf Competition} & {\bf Evaluation Metric} & {\bf Binary} & {\bf Binary Classes} & {\bf Multiclass} & {\bf Multi Classes} & {\bf Dataset} & {\bf Language} \\
\hline
\hline
SemEval \cite{basile:2019} & macro-averaged F1 & 35.0-65.1 &  (non-)hateful & 15.9-57.0 & individual, generic, & Twitter & English \\
& & & & & (non-)aggressive & &\\
SemEval \cite{basile:2019} & macro-averaged F1 & 49.3-73.0 & (non-)hateful & 42.8-70.5 & individual, generic, & Twitter & Spanish \\
& & & & & (non-)aggressive & &\\

GermEval2018 \cite{wiegand-siegel:2018} & macro F1 & 49.0-76.8 & offense, other & 32.1-52.7 & profanity, insult, & Twitter & German \\
& & & & & abuse, other & &\\

GermEval2019 \cite{struss:2019} & macro F1 & 54.9-76.9 & offense, other & 36.8-53.6 & profanity, insult, & Twitter & German\\
& & & & & abuse, other & &\\

GermEval2019 \cite{struss:2019} & macro F1 & 55.4-73.1 & implicit, explicit & & & Twitter & German\\
& & & & & & &\\

GermEval2021 \cite{rischetal:2021} & macro F1 & 36.0-71.8 & toxic comments & & & Facebook & German \\
& & & & & & &\\

GermEval2021 \cite{rischetal:2021} & macro F1 & 61.4-70.0 & engaging comments & & & Facebook & German\\
& & & & & & &\\

GermEval2021 \cite{rischetal:2021} & macro F1 & 59.7-76.3 & fact-claiming & \omit & &Facebook & German\\
& & & comments & & & & \\

Evalita2018 \cite{fersini:2018}& accuracy & 76.5-84.4 & (non-)misogynistic & 29.2-50.1 & stereotype, & Twitter & Italian\\
& & & & & objectification, & &\\
& & & & & dominance, & &\\
& & & & & derailing, & &\\
& & & & & sexual harassment, & &\\
& & & & & discredit, & &\\
& & & & & individual, generic& & \\
& & & & & & &\\
Evalita2018 \cite{fersini:2018} & accuracy & 50.0-70.4 & (non-)misogynistic & 23.2-40.6 & stereotype, & Twitter & English\\
& & & & & objectification, & &\\
& & & & & dominance, & &\\
& & & & & derailing, & &\\
& & & & & sexual harassment, & &\\
& & & & & discredit, & &\\
& & & & & individual, generic & & \\
& & & & & & &\\
HASOC \cite{mandletal:2019} & macro F1 & 50.1-83.1 & (non-)hateful & 45.0-54.5 & hate speech, & Twitter & English\\
\cite{mandletal_2020:2020, mandletal_2021:2021} & & & & & offensive, & &\\
& & & & & profane & &\\
& macro F1 & 45.8-51.1 & (un)targeted &  & & Twitter& English\\
& & & & & & &\\
HASOC \cite{mandletal:2019} & macro F1 & 46.0-81.5 & (non-)hateful & 06.2-58.1 & hate speech, & Twitter & Hindi\\
\cite{mandletal_2020:2020, mandletal_2021:2021} & & & & & offensive, & &\\
& & & & & profane & & \\
& & & & & & &\\
& macro F1 & 49.7-57.5 & (un)targeted &  & & Twitter& Hindi\\
& & & & & & &\\

HASOC \cite{mandletal:2019, mandletal_2020:2020} & macro F1 & 47.9-61.6 & (non-)hateful & 24.0-34.7 & hate speech, & Twitter & German\\
& & & & & offensive, & &\\
& & & & & profane & & \\
& & & & & & &\\
HASOC \cite{mandletal_2021:2021} & macro F1 & 53.9-91.4 & (non-)hateful & & hate speech, & Twitter & Marathi\\
& & & & & offensive, & &\\
& & & & & profane & &\\

\hline
\end{tabular}\end{center}}
\end{center}
\caption{Summary of Competition Results}
\label{table:Table 9}
\end{table*}



\begin{landscape}
\begin{table}
\tiny
\noindent{\begin{center}\begin{spreadtab}{{tabular}{l l l r | r r r | r r r r r r r r r r r | r r}}
\hline
\hline
@\textbf{Wave} & @\textbf{Week} & @\textbf{Lang} & @\textbf{Data} & @\textbf{Examples} & & & @\textbf{Targets} & & & & & & & & & & & @\textbf{Tox} & @\textbf{HS} \\
\hline
& & & & @\textbf{Total} & @\textbf{nonHS} & @\textbf{HS\&Tox} & @\textbf{Sex} & @\textbf{Age} & @\textbf{Gen.} & @\textbf{Rel.} & @\textbf{Nat.} & @\textbf{Imp.} & @\textbf{Stat.} & @\textbf{Pol.} & @\textbf{App.} & @\textbf{Other} & @\textbf{All} & & \\
\hline
\hline
0 & @NGO & @G & @NGO & f3+g3 & 12494 & 366 & 6 & 0 & 0 & 0 & 12 & 0 & 10 & 0 & 0 & 0 & sum(h3:q3) & 338 & g3-s3\\
\hline
1 & 0 & @G & @ON1 & sum(f4:g4) & 24518 & 1147 & 230 & 28 & 34 & 19 & 448 & 0 & 66 & 351 & 0 & 0 & sum(h4:q4) & 92 & g4-s4\\
& 1 & @G & @ON1 & sum(f5:g5) & 21866 & 934 & 192 & 52 & 70 & 52 & 487 & 14 & 49 & 88 & 18 & 98 & sum(h5:q5) & 105 & g5-s5\\
& 2 & @G & @ON1 & sum(f6:g6) & 16019 & 781 & 122 & 41 & 41 & 34 & 282 & 22 & 60 & 103 & 42 & 33 & sum(h6:q6) & 129 & g6-s6\\
& 3 & @G & @ON1 & sum(f7:g7) & 21244 & 1556 & 159 & 75 & 40 & 34 & 567 & 64 & 140 & 211 & 138 & 50 & sum(h7:q7) & 325 & g7-s7\\
& 4 & @G & @ON1 & sum(f8:g8) & 22087 & 1713 & 140 & 45 & 49 & 55 & 600 & 16 & 146 & 305 & 134 & 158 & sum(h8:q8) & 368 & g8-s8\\
& 5 & @G & @ON1 & sum(f9:g9) & 22547 & 1253 & 51 & 61 & 14 & 6 & 303 & 15 & 90 & 82 & 50 & 164 & sum(h9:q9) & 515 & g9-s9\\
& 6 & @G & @ON1 & sum(f10:g10) & 22927 & 3073 & 216 & 133 & 40 & 89 & 1285 & 21 & 330 & 443 & 109 & 125 & sum(h10:q10) & 724 & g10-s10\\
& 7 & @G & @ON1 & sum(f11:g11) & 22007 & 3993 & 344 & 154 & 56 & 80 & 1288 & 34 & 477 & 665 & 155 & 164 & sum(h11:q11) & 1117 & g11-s11\\
& @TN & @G & @ON1 & sum(f12:g12) & 14872 & 128 & 0 & 0 & 0 & 0 & 8 & 0 & 23 & 22 & 0 & 31 & sum(h12:q12) & 55 & g12-s12\\
& 8 & @G & @Tweets & sum(f13:g13) & 25286 & 4512 & 218 & 89 & 25 & 147 & 621 & 79 & 359 & 1416 & 78 & 77 & sum(h13:q13) & 1898 & g13-s13\\
& 9 & @G & @Tweets & f14+g14 & 4487 & 513 & 27 & 10 & 6 & 15 & 74 & 6 & 33 & 181 & 5 & 8 & sum(h14:q14) & 196 & g14-s14\\
\hline
2 & 0 & @FR & @ON1 & sum(f15:g15) & 2780 & 1220 & 115 & 29 & 28 & 33 & 216 & 8 & 59 & 102 & 18 & 1 & sum(h15:q15) & 767 & g15-s15\\
& 1 & @FR & @ON1 & sum(f16:g16) & 7718 & 8460 & 2087 & 270 & 289 & 221 & 2945 & 222 & 693 & 1107 & 502 & 106 & sum(h16:q16) & 2094 & g16-s16\\
& 2 & @FR & @ON1 & sum(f17:g17) & 10066 & 6287 & 1747 & 217 & 188 & 161 & 2210 & 213 & 626 & 1092 & 383 & 50 & sum(h17:q17) & 929 & g17-s17\\
& 3 & @FR & @ON1 & sum(f18:g18) & 9391 & 3609 & 676 & 112 & 104 & 54 & 1059 & 108 & 364 & 596 & 191 & 26 & sum(h18:q18) & 1027 & g18-s18\\
& 4 & @FR & @ON1 & sum(f19:g19) & 4244 & 7757 & 1566 & 106 & 141 & 144 & 4894 & 150 & 1319 & 1229 & 370 & 25 & sum(h19:q19) & 390 & g19-s19\\
& 5 & @FR & @ON1 & sum(f20:g20) & 5375 & 6072 & 1937 & 88 & 130 & 145 & 1969 & 130 & 1103 & 848 & 300 & 71 & sum(h20:q20) & 802 & g20-s20\\
& 6 & @FR & @ON1 & sum(f21:g21) & 4045 & 5006 & 1262 & 74 & 101 & 120 & 1490 & 113 & 1188 & 1057 & 174 & 49 & sum(h21:q21) & 608 & g21-s21\\
& 7 & @FR & @ON1 & sum(f22:g22) & 6827 & 3766 & 608 & 51 & 36 & 62 & 810 & 80 & 850 & 958 & 173 & 91 & sum(h22:q22) & 829 & g22-s22\\
& 8 & @FR & @Tweets & sum(f23:g23) & 8417 & 1576 & 241 & 19 & 33 & 116 & 239 & 55 & 284 & 469 & 64 & 39 & sum(h23:q23) & 519 & g23-s23\\
& 9 & @FR & @Tweets & sum(f24:g24) & 14093 & 2904 & 254 & 35 & 52 & 145 & 275 & 85 & 220 & 844 & 41 & 173 & sum(h24:q24) & 1263 & g24-s24\\
\hline
3 & 1 & @G & @Tweets & sum(f25:g25) & 6854 & 8557 & 666 & 172 & 143 & 468 & 1210 & 104 & 380 & 3108 & 112 & 605 & sum(h25:q25) & 3095 & g25-s25\\
\hline
4 & @1-15 & @G & @Tweets & 33822 & 15395 & 18427 & 718 & 192 & 149 & 222 & 1393 & 86 & 2411 & 9449 & 136 & 3527 & sum(h26:q26) & 3937 & g26-s26\\
\hline
5 & @1-3 & @G & @ON1 & 11304 & 3370 & 7934 & 771 & 110 & 143 & 97 & 1880 & 38 & 528 & 2116 & 170 & 947 & sum(h27:q27) & 2545 & g27-s27\\
& @4-6 & @G & @ON2 & 6267 & 1244 & 5023 & 440 & 78 & 119 & 128 & 875 & 5 & 397 & 2383 & 56 & 1117 & sum(h28:q28) & 1003 & g28-s28\\
& @7-9 & @G & @ON3 & 6042 & 272 & 5770 & 495 & 102 & 169 & 89 & 956 & 15 & 493 & 2418 & 100 & 1571 & sum(h29:q29) & 1145 & g29-s29\\
\hline
\\
\hline
@Total & & & & sum(e3:e29) & sum(f3:f29) & sum(g3:g29) & sum(h3:h29) & sum(i3:i29) & sum(j3:j29) & sum(k3:k29) & sum(l3:l29) & sum(m3:m29) & sum(n3:n29) & sum(o3:o29) & sum(p3:p29) & sum(q3:q29) & sum(r3:r29) & sum(s3:s29) & sum(t3:t29)\\
\hline
\end{spreadtab}\end{center}}
\caption{Results Annotation Process}
\label{table:Table 10}
\end{table}
\end{landscape}



\begin{table*}
\small
\begin{center}
\noindent{\begin{center}\begin{tabular}{l l l r r r r r r}
\hline
\hline
 & & & \textbf{Base} & & & \textbf{Tuned} & & \\
\hline
\textbf{Lang} & \textbf{Dataset} & \textbf{Model} & \textbf{Precision} & \textbf{Recall} & \textbf{F1} & \textbf{Precision} & \textbf{Recall} & \textbf{F1}\\
\hline
\hline
G & Comments & GELECTRA\textsubscript{Base} & 44.4 & 47.4 & 39.3 & 76.0 & 75.5 & 75.4\\ 
G & Comments & GBERT\textsubscript{Base} & 44.1 & 49.8 & 33.7 & 75.7 & 75.4 & 75.4\\
 \\
G & Comments & mBERT\textsubscript{Base} & 25.0 & 50.0 & 33.3 & 74.7 & 74.4 & 74.4\\
G & Comments & XLM-RoBERTa\textsubscript{Base} & 25.0 & 50.0 & 33.3 & 75.7 & 75.0 & 74.9\\
 \\
G & Comments & Twitter-XLM-RoBERTa\textsubscript{Base} & 41.7 & 50.0 & 33.3 & 75.4 & 75.1 & 75.0\\
\hline
FR & Comments & flauBERT\textsubscript{Base} & 49.6 & 49.6 & 49.3 & 66.5 & 66.5 & 66.4\\
FR & Comments & french RoBERTa\textsubscript{Base} & 46.7 & 49.1 & 37.9 & 62.3 & 62.3 & 62.3\\ 
 \\
FR & Comments & mBERT\textsubscript{Base} & 58.0 & 50.0 & 33.4 & 67.4 & 67.2 & 66.9\\
FR & Comments & XLM-RoBERTa\textsubscript{Base} & 49.5 & 49.9 & 37.3 & 67.2 & 67.2 & 67.1\\
 \\
FR & Comments & Twitter-XLM-RoBERTa\textsubscript{Base} & 37.4 & 50.0 & 33.3 & 67.2 & 67.1 & 67.0\\
\hline
G \& FR & Comments & mBERT\textsubscript{Base} & 40.3 & 49.9 & 33.4 & 79.8 & 79.6 & 79.6\\
G \& FR & Comments & XLM-RoBERTa\textsubscript{Base} & 50.5 & 50.5 & 50.0 & 79.9 & 79.9 & 79.8\\
\\
G \& FR & Comments & Twitter-XLM-RoBERTa\textsubscript{Base} & 50.5 & 50.5 & 50.0 & 79.9 & 79.9 & 79.8\\
\hline
\hline
G & Tweets & GELECTRA\textsubscript{Base} & 52.7 & 51.0 & 42.1 & 79.5 & 79.3 & 79.3\\ 
G & Tweets & GBERT\textsubscript{Base} & 56.0 & 55.6 & 54.9 & 78.8 & 78.8 & 78.8\\
 \\
G & Tweets & mBERT\textsubscript{Base} & 25.0 & 50.0 & 33.3 & 78.2 & 78.2 & 78.2\\
G & Tweets & XLM-RoBERTa\textsubscript{Base} & 61.4 & 50.0 & 33.4 & 78.6 & 78.3 & 78.3\\
 \\
G & Tweets & Twitter-XLM-RoBERTa\textsubscript{Base} & 50.0 & 50.0 & 33.3 & 78.7 & 78.7 & 78.7\\
\hline
FR & Tweets & flauBERT\textsubscript{Base} & 51.4 & 51.4 & 51.3 & OF & OF & OF\\
FR & Tweets & french RoBERTa\textsubscript{Base} & 58.4 & 50.2 & 34.0 & OF & OF & OF\\ 
 \\
FR & Tweets & mBERT\textsubscript{Base} & 55.0 & 50.0 & 33.5 & OF & OF & OF\\
FR & Tweets & XLM-RoBERTa\textsubscript{Base} & 25.0 & 50.0 & 33.3 & OF & OF & OF\\
 \\
FR & Tweets & Twitter-XLM-RoBERTa\textsubscript{Base} & 25.0 & 50.0 & 33.3 & OF & OF & OF\\
\hline
G \& FR & Tweets & mBERT\textsubscript{Base} & 52.7 & 51.7 & 46.8 & 76.4 & 76.4 & 76.4\\
G \& FR & Tweets & XLM-RoBERTa\textsubscript{Base} & 25.0 & 50.0 & 33.3 & 78.1 & 77.9 & 77.9\\
\\
G \& FR & Tweets & Twitter-XLM-RoBERTa\textsubscript{Base} & 25.0 & 50.0 & 33.3 & 77.8 & 77.4 & 77.4\\
\hline
\hline
G & Comments \& Tweets & GELECTRA\textsubscript{Base} & 54.3 & 52.7 & 47.7 & 76.5 & 76.4 & 76.4\\ 
G & Comments \& Tweets & GBERT\textsubscript{Base} & 46.5 & 47.7 & 42.8 & 76.6 & 76.5 & 76.5\\
 \\
G & Comments \& Tweets & mBERT\textsubscript{Base} & 50.5 & 50.3 & 44.1 & 74.0 & 73.9 & 73.9\\
G & Comments \& Tweets & XLM-RoBERTa\textsubscript{Base} & 52.5 & 51.3 & 44.7 & 75.2 & 75.2 & 75.2\\
 \\
G & Comments \& Tweets & Twitter-XLM-RoBERTa\textsubscript{Base} & 57.9 & 50.0 & 33.6 & 75.4 & 75.3 & 75.2\\
\hline
FR & Comments \& Tweets & flauBERT\textsubscript{Base} & 48.2 & 48.4 & 46.5 & 67.6 & 67.1 & 66.9\\
FR & Comments \& Tweets & french RoBERTa\textsubscript{Base} & 25.0 & 50.0 & 33.3 & 62.6 & 62.6 & 62.6\\ 
 \\
FR & Comments \& Tweets & mBERT\textsubscript{Base} & 51.6 & 50.0 & 33.8 & 67.3 & 67.3 & 67.3\\ 
FR & Comments \& Tweets & XLM-RoBERTa\textsubscript{Base} & 50.0 & 50.0 & 33.3 & 67.4 & 67.3 & 67.3\\
 \\
FR & Comments \& Tweets & Twitter-XLM-RoBERTa\textsubscript{Base} & 49.0 & 49.8 & 36.8 & 68.2 & 68.0 & 68.0\\
\hline
G \& FR & Comments \& Tweets & mBERT\textsubscript{Base} & 25.0 & 50.0 & 33.3 & 79.3 & 79.2 & 79.2\\
G \& FR & Comments \& Tweets & XLM-RoBERTa\textsubscript{Base} & 25.0 & 50.0 & 33.3 & 71.2 & 71.1 & 71.1\\
\\
G \& FR & Comments \& Tweets & Twitter-XLM-RoBERTa\textsubscript{Base} & 25.0 & 50.0 & 33.3 & 71.2 & 71.1 & 71.1\\
\hline
\end{tabular}\end{center}}
\end{center}
\caption{BERT-Models Final Results}
\label{table:Table 11}
\end{table*}


\begin{landscape}
\begin{table}
\small
\noindent{\begin{center}\begin{spreadtab}{{tabular}{l l | r r r | r r r r r r r r r r r | r r}}
\hline
\hline
@\textbf{Lang} & @\textbf{Data} & @\textbf{Examples} & & & @\textbf{Targets} & & & & & & & & & & & @\textbf{Tox} & @\textbf{HS} \\
\hline
& & @\textbf{Total} & @\textbf{nonHS} & @\textbf{HS\&Tox} & @\textbf{Sex} & @\textbf{Age} & @\textbf{Gender} & @\textbf{Rel.} & @\textbf{Nat.} & @\textbf{Imp.} & @\textbf{Status} & @\textbf{Pol.} & @\textbf{App.} & @\textbf{Other} & @\textbf{All} & & \\
\hline
\hline
@G & @Comments & 225232 & 201445 & 23787 & 2288 & 751 & 574 & 535 & 7032 & 206 & 2098 & 5846 & 813 & 2609 & sum(f3:o3) & 5629 & e3-q3\\
@G & @Tweets & 83851 & 51885 & 31966 & 1628 & 463 & 324 & 848 & 3295 & 275 & 3179 & 14145 & 331 & 4213 & sum(f4:o4) & 9107 & e4-q4\\
@FR & @Comments & 86213 & 47307 & 38906  & 9062 & 890 & 976 & 862 & 14737 & 955 & 5492 & 6424 & 1961 & 389 & sum(f5:o5) & 6910 & e5-q5\\
@FR & @Tweets & 26750 & 22333 & 4417 & 492 & 54 & 83 & 257 & 511 & 140 & 493 & 1302 & 105 & 212 & sum(f6:o6) & 1749 & e6-q6\\
\hline
@Total & @Both & sum(c3:c6) & sum(d3:d6) & sum(e3:e6) & sum(f3:f6) & sum(g3:g6) & sum(h3:h6) & sum(i3:i6) & sum(j3:j6) & sum(k3:k6) & sum(l3:l6) & sum(m3:m6) & sum(n3:n6) & sum(o3:o6) & sum(f7:o7) & sum(q3:q6) & e7-q7\\
\hline
\end{spreadtab}\end{center}}
\caption{Unique Annotations in the Swiss Hate Speech Corpus}
\label{table:Table 12}
\end{table}
\end{landscape}


\sisetup{
  round-mode = places, 
  round-precision = 0, 
}

\begin{landscape}
\begin{table}
\small
\noindent{\begin{center}\begin{spreadtab}{{tabular}{l | r | S | r | S | r | S | r | S}}
\hline
\hline
& \multicolumn{4}{c}{\textbf{@ONs}} & \multicolumn{4}{|c}{\textbf{@Twitter}} \\
\hline
& \multicolumn{1}{c}{@G HS =} & 18158 & \multicolumn{1}{|c}{@FR HS =} & 31996 & \multicolumn{1}{|c}{@G HS =} & 22859 & \multicolumn{1}{|c}{@FR HS =} & 2668 \\
\hline
@\textbf{Target} & @HSGroup & \% & @HSGroup & \% & @HSGroup & \% & @HSGroup & \% \\
\hline
\hline
@Sex & 2288 & (b4/c2)*100 & 9062 & (d4/e2)*100 & 1628 & (f4/g2)*100 & 492 & (h4/i2)*100 \\
@Age & 751 & (b5/c2)*100 & 890 & (d5/e2)*100 & 463 & (f5/g2)*100 & 54 & (h5/i2)*100 \\
@Gender & 574 & (b6/c2)*100 & 976 & (d6/e2)*100 & 324 & (f6/g2)*100 & 83 & (h6/i2)*100 \\
@Rel. & 535 & (b7/c2)*100 & 862 & (d7/e2)*100 & 848 & (f7/g2)*100 & 257 & (h7/i2)*100 \\
@Nat. & 7032 & (b8/c2)*100 & 14737 & (d8/e2)*100 & 3295 & (f8/g2)*100 & 511 & (h8/i2)*100 \\
@Imp. & 206 & (b9/c2)*100 & 955 & (d9/e2)*100 & 275 & (f9/g2)*100 & 140 & (h9/i2)*100 \\
@Status & 2098 & (b10/c2)*100 & 5492 & (d10/e2)*100 & 3179 & (f10/g2)*100 & 493 & (h10/i2)*100 \\
@Pol. & 5846 & (b11/c2)*100 & 6424 & (d11/e2)*100 & 14145 & (f11/g2)*100 & 1302 & (h11/i2)*100 \\
@App. & 813 & (b12/c2)*100 & 1961 & (d12/e2)*100 & 331 & (f12/g2)*100 & 105 & (h12/i2)*100 \\
@Other & 2609 & (b13/c2)*100 & 389 & (d13/e2)*100 & 4213 & (f13/g2)*100 & 212 & (h13/i2)*100 \\
\hline
\end{spreadtab}\end{center}}
\caption{Percentages of Hate Speech per Multilabel Target Group}
\label{table:Table 13}
\end{table}
\end{landscape}


\end{document}